\documentclass[journal]{IEEEtran}
\usepackage{graphicx}
\usepackage{epsfig}
\usepackage{latexsym}
\usepackage{amsfonts}
\usepackage{here}
\usepackage{rawfonts}
\usepackage[latin1]{inputenc}
\usepackage[T1]{fontenc}
\usepackage{calc}
\usepackage{url}
\usepackage{enumerate}
\usepackage{color}
\usepackage[tbtags]{amsmath}
\usepackage{amssymb}
\usepackage{upref}
\usepackage{epic,eepic}
\usepackage{times}
\usepackage{dsfont}
\usepackage{comment}
\usepackage{cite}
\usepackage{subcaption}
\usepackage{graphicx}
\usepackage[font={small}]{caption}
\usepackage[font={small}]{subcaption}

\usepackage{multirow}
\usepackage{booktabs}

\usepackage{stfloats}
\usepackage{mathtools}
\usepackage{amsmath}

\usepackage{algorithm}
\usepackage{algpseudocode}

\newtheorem{theorem}{\bf{Theorem}}

\newtheorem{corollary}{\bf{Corollary}}

\newtheorem{remark}{Remark}

\ifCLASSINFOpdf
\else
\fi
\begin{document}

	\title{A Generalized Hierarchical Federated Learning Framework with Theoretical Guarantees}
	
	\author{{Seyed~Mohammad~Azimi-Abarghouyi,~\IEEEmembership{Member,~IEEE},~and~Carlo~Fischione,~\IEEEmembership{Fellow,~IEEE}\vspace{0pt}}

		\thanks{The authors are with the School of Electrical Engineering and Computer Science, KTH Royal Institute of Technology, Stockholm, Sweden (Emails: $\bigl\{$seyaa,\ carlofi$\bigr\}$@kth.se). This project has received funding from the European Union's Horizon Europe programme under grant agreement No. 101137954. The authors would like to thank the rest of the BATTwin consortium for supporting this research.}
	}

	\maketitle
	
	\vspace{-15pt}
	\begin{abstract}
		Almost all existing hierarchical federated learning (FL) models are limited to two aggregation layers, restricting scalability and flexibility in complex, large-scale networks. In this work, we propose a Multi-Layer Hierarchical Federated Learning framework ({\fontfamily{lmtt}\selectfont QMLHFL}), which appears to be the first study that generalizes hierarchical FL to arbitrary numbers of layers and network architectures through nested aggregation, while employing a layer-specific quantization scheme to meet communication constraints. We develop a comprehensive convergence analysis for {\fontfamily{lmtt}\selectfont QMLHFL} and derive a general convergence condition and rate that reveal the effects of key factors, including quantization parameters, hierarchical architecture, and intra-layer iteration counts. Furthermore, we determine the optimal number of intra-layer iterations to maximize the convergence rate while meeting a deadline constraint that accounts for both communication and computation times. Our results show that {\fontfamily{lmtt}\selectfont QMLHFL} consistently achieves high learning accuracy, even under high data heterogeneity, and delivers notably improved performance when optimized, compared to using randomly selected values.
	\end{abstract}
	\vspace{0pt}
	\begin{IEEEkeywords}
		Machine learning, federated learning, hierarchical systems, quantization
	\end{IEEEkeywords}
	\vspace{-5pt}
	\section{Introduction}
	Federated Learning (FL) has gained significant attention as an approach that enables distributed model training without requiring centralized access to raw data~\cite{mcmahan}. This method is especially valuable in situations where data is costly to gather or difficult to consolidate. Additionally, FL supports improved computational efficiency by enabling simultaneous model training across multiple devices. 
	
	FL was originally developed for a single centralized server~\cite{mcmahan}. More recently, hierarchical FL has emerged, incorporating both cloud and edge servers~\cite{letaief, castiglia}. This model typically involves two layers of aggregation: (1) edge aggregation, performed at edge servers that group devices, and (2) cloud aggregation, conducted at a cloud server to merge results from multiple edge servers. Despite the benefits of hierarchical FL, existing frameworks and theoretical guarantees remain limited to two-layer designs, reducing their flexibility and scalability. This constraint becomes particularly problematic in scenarios where multiple intermediary layers are essential for optimizing communication and computation efficiency, handling large-scale geographic distributions of devices, and ensuring efficient data processing. In many real-world scenarios, networks exhibit multi-layer hierarchical architectures~\cite{analu, harpreet}. For example, cloud architectures distribute workloads across local, regional, national, and global tiers; IoT infrastructures aggregate data through device, gateway, fog, and cloud layers; and cellular networks rely on femtocell, picocell, microcell, macrocell, and core network tiers. Similarly, healthcare systems organize multi-layered data sharing across individuals, hospitals, regional health organizations, and national and global providers.
	
	Such large-scale hierarchical architectures involve many nodes across different layers, each subject to distinct communication and bandwidth constraints. In such settings, effective deployment of quantization is essential for managing communication cost, as it has long been recognized as a critical component of communication-efficient FL systems~\cite{alish, pedar}.
	
	To address these, in this work, we propose a quantization-aware \textit{Multi-Layer Hierarchical Federated Learning} framework, called {\fontfamily{lmtt}\selectfont QMLHFL}, which extends hierarchical FL to accommodate any number of layers and is supported by theoretical guarantees. It introduces a new dimension of hierarchical depth into FL research---an aspect that was previously fixed and unexplored due to the prevailing limitation to at most two layers.
	\vspace{-5pt}
	\subsection{Why Multi-Layer Hierarchical FL is Necessary}
	\begin{itemize}
		\item \textbf{Enhanced Scalability and Adaptability}:
			{\fontfamily{lmtt}\selectfont QMLHFL} offers a unified framework that enables FL to operate over arbitrary hierarchical architectures, adapting to diverse network topologies without the constraints of a rigid design, and ensuring scalable operation across networks of any size and region.
		\item \textbf{Improved Communication and Energy Efficiency}:  
			By enabling localized aggregations and distributing communication efficiently across multiple layers, {\fontfamily{lmtt}\selectfont QMLHFL} reduces the communication burden on central servers, optimizes bandwidth usage, and lowers latency by bringing devices closer to the edge. This hierarchical structure further limits long-range inter-layer transmissions and their associated congestion, thereby enhancing overall energy efficiency and resource utilization.
		\item \textbf{Handling Device and Network Heterogeneity}:  
			The ability of {\fontfamily{lmtt}\selectfont QMLHFL} to distribute learning across multiple layers helps accommodate varying computational and communication capacities among devices and servers, improving convergence stability under heterogeneous conditions.
		\item \textbf{Increased Robustness and Privacy}:  
			By distributing aggregation responsibilities across multiple layers, {\fontfamily{lmtt}\selectfont QMLHFL} reduces the risk of single points of failure and isolates potential malicious behavior within local clusters. It restricts communication to short-range or administrative domains and applies privacy-preserving and integrity-check mechanisms at each layer, thereby ensuring that data exchange remains confined within trusted domains---further strengthening both privacy and security.
		\item \textbf{Position within the FL Spectrum}:  
			Centralized and decentralized FL represent two extremes of the design spectrum~\cite{advanceFL}: the former ensures global coordination but faces scalability and communication bottlenecks, while the latter promotes autonomy but relies on dense and stable peer-to-peer connectivity supported by a well-established communication graph. Maintaining such connectivity across large, heterogeneous, or geographically dispersed networks is impractical due to latency, routing complexity, link instability, and inherent physical and interoperability constraints. Moreover, decentralized FL lacks structured trust boundaries, complicating privacy enforcement, and often suffers from model inconsistency and biased convergence due to partial device participation.  
			{\fontfamily{lmtt}\selectfont QMLHFL} provides a balanced and broad middle ground by introducing structured aggregation layers that mirror real-world communication hierarchies. In doing so, it can practically interconnect fundamentally different environments, domains, or isolated clusters through intermediate layers that standardize and bridge their communications, which would be infeasible under purely decentralized FL. This organization localizes coordination while inclusively engaging all devices in the global learning process, ensuring that every participant both contributes to and benefits from the collective model. Consequently, {\fontfamily{lmtt}\selectfont QMLHFL} combines the scalability of decentralized designs with the reliability, fairness, coordination, and privacy protection of centralized FL, enabling efficient and inclusive learning in large-scale, heterogeneous systems.
	\end{itemize}
	
	\vspace{-5pt}
	\subsection{Related Works}
	Hierarchical FL with quantization has been studied in~\cite{letaief, ourotherjournal}, where the convergence behavior and optimal aggregation intervals under quantized communication have been analyzed to improve communication efficiency and learning performance. The integration of hierarchical FL with over-the-air computation has been examined in~\cite{ourjournal, ourconf, hier_ota}, which demonstrated that analog aggregation and hierarchical clustering can significantly enhance scalability and robustness against interference and data heterogeneity in wireless networks. An alternative approach to alleviating communication limitations focuses on model pruning~\cite{prun2}, where redundant parameters are removed to reduce communication cost while maintaining model accuracy. Delay-aware frameworks are also proposed in~\cite{purdue, chat}. In~\cite{purdue}, a local-global synchronization mechanism is introduced to improve learning efficiency under delay, while~\cite{chat} analyzes hierarchical FL with group-based local averaging and periodic synchronization governed by delay-dependent timing. Furthermore, several network optimization problems have been addressed in~\cite{lim, bennis, tony, energy_resource, feduc, place, wu, context, scheduling, wen, cost, poor, reen}. For instance, resource allocation and incentive design for hierarchical FL have been formulated in~\cite{lim}, while~\cite{bennis, tony} investigated joint user association and resource allocation strategies to improve convergence efficiency and reduce latency. Energy-efficient designs have been introduced in~\cite{energy_resource, reen}, which jointly optimize energy transfer, aggregation frequency, and edge association to balance learning accuracy and power consumption. Clustering and aggregator placement strategies have been investigated in~\cite{feduc, place} to improve scalability and reduce implementation cost, while heterogeneity-aware client association and staleness control have been developed in~\cite{wu} to enhance convergence under practical network conditions. Context-aware and scheduling-based frameworks have been proposed in~\cite{context, scheduling, wen} to select active devices and allocate communication resources dynamically, enabling more adaptive and stable training processes. Finally, fairness- and cost-oriented optimization~\cite{cost} and graph-based joint clustering and resource allocation methods~\cite{poor} have been explored to improve convergence performance in heterogeneous and resource-constrained networks. 
		These works build upon the traditional FL paradigm, in which model or gradient parameters are exchanged and aggregated across both lower and higher layers using gradient descent-based optimization. Beyond gradient descent, a hierarchical FL approach leveraging the Alternating Direction Method of Multipliers (ADMM) has been proposed in~\cite{azimi3}. However, a fundamental assumption of all of these works is that they are limited to two-layer architectures. To the best of our knowledge, \cite{muulti} is the only work that considers multi-layer hierarchical architectures for FL. While it represents a valuable first approach in this direction, it does not provide a mathematical model of the problem or offer any theoretical guarantees.
	\vspace{-5pt}
	\subsection{Contributions}
	\textit{General Framework:} We propose a scalable and flexible learning algorithm that efficiently aggregates models from lower to higher layers using a nested approach, ultimately constructing a global model at the cloud server. Our approach leverages layer-specific iteration counts, allowing each layer to adapt independently. Additionally, we develop a quantization scheme for the transmission at each layer, highlighting the distinct communication capacities of each layer.

	\textit{Convergence Analysis:} We develop a comprehensive convergence analysis of the proposed algorithm under arbitrary layer configurations. Furthermore, we derive tractable expressions for the convergence condition and convergence rate, and offer design insights supported by analyses of special cases. Notably, two-layer hierarchical analysis is merely a special case of our generalized approach. Existing two-layer analysis does not capture the full complexity and interactions present in deeper hierarchies, omitting many critical terms that our approach incorporates. Our new analysis provides guarantees that the convergence speed of the proposed algorithm scales with the product of the intra-layer iteration counts, highlighting the full exploitation of the multi-layer potential within the framework. Additionally, while the convergence rate is a function of the total number of devices and the number of edge servers at each layer, the convergence condition depends on the full set of architectural parameters.
	
	\textit{Learning Optimization:} Building on our analysis, we formulate an optimization problem to determine the optimal number of intra-layer iterations for each layer, with the objective of maximizing the convergence rate while satisfying a completion deadline. The optimal values strike a balance between convergence speed and post-convergence error, while accounting for both computation and communication times over the network. We then propose an algorithm based on geometric programming to solve this nonlinear problem effectively. Notably, despite its importance, such an optimization has not been explored even in the traditional two-layer hierarchical FL.
	
	\textit{Insights:} The experimental results demonstrate strong convergence and learning performance across various hierarchical depths, effectively managing data heterogeneity. In addition, the proposed optimization yields a notable improvement in performance. Furthermore, increasing the number of layers leads to markedly faster run-time to convergence. Analytical results reveal that, under a fixed number of devices, increasing the number of edge servers in a layer amplifies post-convergence error. Furthermore, quantization effects are more pronounced in lower layers, especially at the device layer. Also, in computation-limited scenarios, directing all iterations to the highest layer yields optimal convergence when quantization is absent.

	\vspace{-5pt}
	\section{System Model}
	In this section, we present a general hierarchical setup with multiple layers and formulate the corresponding distributed learning problem.
	
	\begin{table}[t]
			\centering
			\caption{List of Key Notations}
			\renewcommand{\arraystretch}{1.2}
			\setlength{\tabcolsep}{4pt}
			\begin{tabular}{ll}
				\toprule
				\textbf{Notation} & \textbf{Description} \\ 
				\midrule
				$N$ & Number of aggregation layers \\
				$N_\text{tot}$ & Total number of devices in the network \\ $b$ & Mini-batch size\\ $D_i$ & Number of local training samples at device $i$ \\
				$\mathcal{C}_n$ & Set of edge servers at layer $n$, with cardinality $C_n = |\mathcal{C}_n|$ \\
				$\mathcal{C}_{n}^{i_n}$ & Set of nodes in layer $n-1$ connected to server $i_n$ \\
				$|{\cal C}_{n}^{i_n}|$ & Number of devices under server $i_n$ \\
				$\tau_n$ & Intra-layer iteration count at layer $n$ \\
				$\mu$ & Learning rate \\
				$q_n$ & Variance of quantization error at layer $n$ \\ $f_i$ & CPU cycle frequency of device $i$
				\\
				$t_\text{CP}$ & Maximum local computation time \\
				$t_\text{DE}$ & Transmission time between device and its edge server \\
				$t_{\text{E}_{n-1,n}}$ & Transmission time between edge layers $n{-}1$ and $n$ \\
				$T_\text{d}$ & Completion deadline \\
				$\alpha$ & Trade-off coefficient in learning optimization \\
				$L$ & Lipschitz constant of the loss function gradient \\
				$\sigma^2$ & Upper bound on stochastic gradient variance \\
				\bottomrule
		\end{tabular}
		\label{tab:notations}
	\end{table}
	\vspace{-5pt}
	\subsection{Multi-Layer Setup}
	Consider an $N$-layer hierarchical system comprising a total of $N_\text{tot}$ devices, each identified by an index $i$ belonging to a set $\mathcal{C}$. The devices constitute layer $0$, and the system is structured with $N-1$ layers of edge servers above them, culminating in a cloud server at the $N$-th layer.
	
	In the $n$-th layer, we denote the set of edge servers as $\mathcal{C}_n$ with the cardinality $C_n$. Additionally, for a given edge server with index $i_n$ belonging to $\mathcal{C}_n$, we identify a set of edge servers in the previous layer, $n-1$, that are connected to it. This set is denoted as $\mathcal{C}_{n}^{i_n}$. The total count of devices connected to those edge servers in the first layer, which are hierarchically linked to any of the edge servers in $\mathcal{C}_{n}^{i_n}$, is represented as $|\mathcal{C}_{n}^{i_n}|$. This device count corresponds to the cardinality of the device set as $\left\{i \in \mathcal{C}_1^{i_1}, i_1 \in \mathcal{C}_2^{i_2}, \ldots, i_{n-1} \in \mathcal{C}_n^{i_n}\right\}$, which equals $|{\cal C}_{n}^{i_{n}}| 
		= \sum_{i \in \mathcal{C}_1^{i_1}, \ldots, i_{n-1} \in \mathcal{C}_{n}^{i_{n}}} 1$. A representative example is shown in Fig. 1.
	\vspace{-5pt}
	\subsection{Learning Problem}
	Each device $i$ owns a local dataset ${\cal D}_i$ of size $D_i$. There
		is no prior knowledge of local data statistics of the devices.
		The learning model is parameterized by the vector $\mathbf{w} \in \mathbb{R}^{d}$, with $d$ denoting the number of model parameters. 
		The local loss at device $i$ is defined as
		\begin{equation}
			F_i(\mathbf{w}) = \frac{1}{D_i} \sum_{(\mathbf{x},y) \in {\cal D}_i} \ell(\mathbf{x},y;\mathbf{w}),
		\end{equation}
		where $\ell(\mathbf{x},y;\mathbf{w})$ is the loss evaluated on sample $(\mathbf{x},y)$ using parameter vector $\mathbf{w}$. 
		Accordingly, the global loss function across all $N_\text{tot}$ devices is given by
		\begin{align}
			\label{main_FL}
			F(\mathbf{w}) = \frac{1}{N_\text{tot}} \sum_{i=1}^{N_\text{tot}} F_i(\mathbf{w}).
		\end{align}
		Equivalently, the same global loss in~\eqref{main_FL} can be expressed using the hierarchical setup. 
		Specifically, 
		for an edge server $i_1 \in \mathcal{C}_1$ at layer~1, its aggregated loss is
		\begin{equation}
			F_{i_1}(\mathbf{w}) = \frac{1}{|{\cal C}_{1}^{i_1}|} \sum_{i \in \mathcal{C}_1^{i_1}} F_i(\mathbf{w}).
		\end{equation}
		More generally, for a server $i_n \in \mathcal{C}_n$ at layer $n$, the aggregated loss is defined as
		\begin{equation}
			F_{i_n}(\mathbf{w}) = \sum_{i_{n-1} \in \mathcal{C}_n^{i_n}} 
			\frac{|{\cal C}_{n-1}^{i_{n-1}}|}{|{\cal C}_{n}^{i_n}|} F_{i_{n-1}}(\mathbf{w}).
		\end{equation}
		At the top layer ($N$), the cloud server computes the global loss as
		\begin{align}
			\label{lossfunction_hier}
			F(\mathbf{w}) = \frac{1}{N_\text{tot}} 
			\sum_{i_{N-1} \in \mathcal{C}_N} |{\cal C}_{N-1}^{i_{N-1}}| \, F_{i_{N-1}}(\mathbf{w}).
		\end{align}
		Then, the objective of the learning process is to find the optimal parameter vector as
		\begin{align}
			\label{objective}
			\mathbf{w}^* = \arg\min_{\mathbf{w}} F(\mathbf{w}).
	\end{align}
	In the following section, we present our algorithm designed to solve~\eqref{objective} while accounting for communication constraints.
	\begin{figure}[tb!]
		\hspace{0pt}
		\includegraphics[width =3.4in]{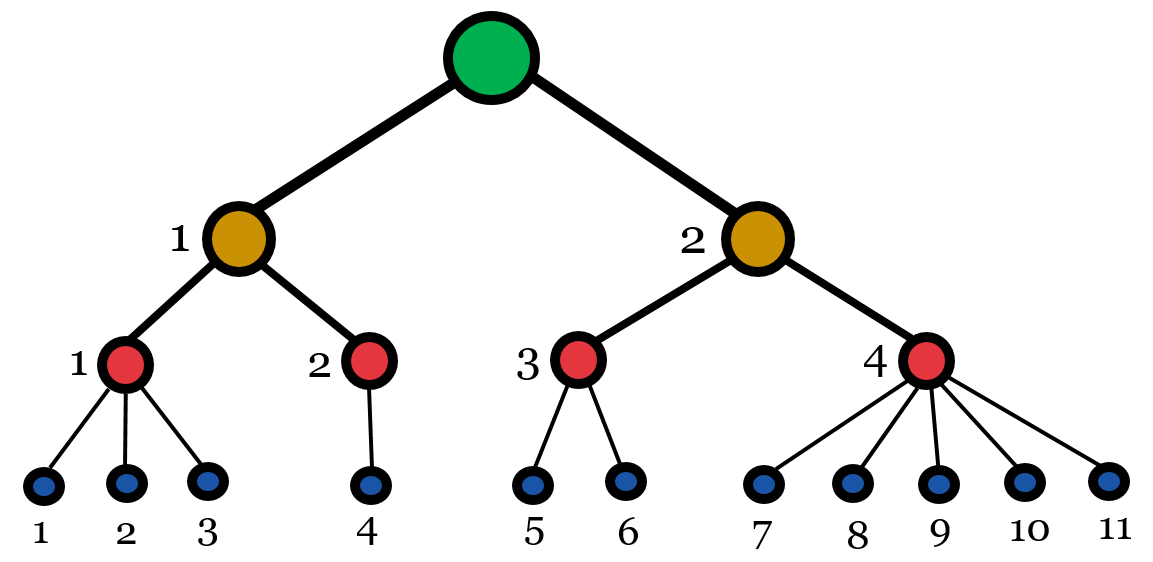}
		\vspace{-4pt}
		\caption{An example of a 3-layer hierarchical system where the cloud server is represented by a green node, edge servers at layer 2 are shown as orange nodes, edge servers at layer 1 as red nodes, and devices as blue nodes. Node sizes increase with hierarchy level. In this system, $N_\text{tot} = 11$, $C_1 = 4$, $C_2 = 2$, $|{\cal C}_1^{1}| = 3$, $|{\cal C}_1^{2}| = 1$, $|{\cal C}_1^{3}| = 2$, $|{\cal C}_1^{4}| = 5$, $|{\cal C}_2^{1}| = 4$, and $|{\cal C}_2^{2}| = 7$.}
		\vspace{-5pt}
	\end{figure}

	\section{Multi-Layer Hierarchical Federated Learning}
	{\fontfamily{lmtt}\selectfont QMLHFL} for \( N \) aggregation layers, called $N$-layer {\fontfamily{lmtt}\selectfont QMLHFL}, operates through three key stages in each global iteration $t$: local update, nested edge aggregation, and cloud aggregation. This structure enables scalable distributed learning through a temporal update process across layers. Let \( t_1, t_2, \dots, t_N \) denote the intra-layer iteration indices, where \( t_1 \) corresponds to the local update at devices, constrained by a maximum count of \( \tau_1 \), and \( t_n \) (\( 2 \leq n \leq N \)) represents updates at edge layer \( n-1 \), with a maximum of \( \tau_n \). 
	
	Each layer employs a distinct quantization function for its transmission to the next layer, tailored to its communication constraint, such as bandwidth availability. For example, devices typically operate with lower bandwidth in a shared wireless medium, whereas edge servers, connected to higher layers via backhaul links, benefit from greater communication resources.
	
	\vspace{-5pt}
	\subsection{Algorithm}
	\textbf{Local Update:} 
	Each device $i$ updates its local model based on stochastic gradient descent (SGD) through multiple iterations $t_1 \in \left\{1,\ldots,\tau_1\right\}$:
	\begin{align}
		\label{local_update}
		\mathbf{w}_{t_1+1,t_2,\ldots,t_{N}}^{i,t} = \mathbf{w}_{t_1,\ldots,t_{N}}^{i,t} - \mu \nabla F_{i}(\mathbf{w}_{t_1,\ldots,t_{N}}^{i,t},\boldsymbol\xi_{t_1,\ldots,t_{N}}^{i,t}), \nonumber\\\forall i \in {\cal C},
	\end{align}
	where $\boldsymbol\xi_{t_1,\ldots,t_{N}}^{i,t} \in {\cal D}_i$ denotes a selected local training mini-batch of fixed size $b$\footnote{The mini-batch size is selected to ensure that each device can complete its local computation before the deadline required for synchronous transmission.} and $\mu$ is the learning rate. In expectation, this step coincides with a gradient descent update on the local loss $F_i(\mathbf{w})$.
	
	\textbf{Nested Edge Aggregation at Layer $n$:} Each edge server $i_n$ at layer $n$ aggregates updates from its respective child nodes (devices or edge servers in the previous layer $n-1$) through multiple iterations $t_{n+1} \in \left\{1,\ldots,\tau_{n+1}\right\}$:
	\begin{align}
		\label{edge_update}
		&\mathbf{v}_{ t_{n+1}+1,t_{n+2},\ldots,t_{N}}^{i_n,t} = \mathbf{v}_{ t_{n+1},\ldots,t_{N}}^{i_n,t} + \sum_{i_{n-1}\in {\cal C}_{n}^{i_n}}^{} \frac{|{\cal C}_{n-1}^{i_{n-1}}|}{|{\cal C}_{n}^{i_n}|}\nonumber\\&Q_{n}(\mathbf{v}_{\tau_{n}, t_{n+1},\ldots,t_{N}}^{i_{n-1},t} - \mathbf{v}_{ t_{n+1},\ldots,t_{N}}^{i_n,t}), \forall i_n \in {\cal C}_{n},
	\end{align}
	where \( Q_n(\cdot) \) denotes the quantization function applied to the model updates computed at layer \( n{-}1 \) after \( \tau_n \) iterations, i.e., \( \mathbf{v}_{\tau_{n}, t_{n+1},\ldots,t_{N}}^{i_{n-1},t}- \mathbf{v}_{ t_{n+1},\ldots,t_{N}}^{i_n,t}, \forall i_{n-1} \), for transmission to layer \( n \). In particular, \( Q_1(\cdot) \) corresponds to the quantization function used at the device layer. The weighting $\tfrac{|{\cal C}_{n-1}^{i_{n-1}}|}{|{\cal C}_{n}^{i_n}|}$ matches the recursive definition of $F_{i_n}(\mathbf{w})$, so this aggregation step approximates a gradient descent update on $F_{i_n}(\mathbf{w})$.
	
	Then, the updates are propagated back to ensure consistency as
	\begin{align}
		\label{edge_broadcast}
		&\mathbf{w}_{0,\ldots,0,t_n+1, t_{n+1},\cdots,t_{N}}^{i,t} = \ldots =
		\mathbf{v}_{0, t_n+1,t_{n+1},\ldots,t_{N}}^{i_{n-1},t} \nonumber\\
		&= \mathbf{v}_{t_n+1,t_{n+1},\ldots,t_{N}}^{i_{n},t}, \quad \forall i \in {\cal C}_1^{i_1}, \ldots,\forall i_{n-1}\in {\cal C}_{n}^{i_n}.
	\end{align}
	This process exhibits a nested structure, where each node aggregates updates that already represent combined contributions from all nodes below it in the hierarchy. Rather than aggregating raw updates, each layer fuses information that has been progressively distilled through prior aggregations at lower layers. The term \textit{nested} here refers to a recursive, self-similar organization, where the same type of operation repeats within each layer---much like a nested lattice constellation~\cite{lattice}, where zooming into any part reveals the same underlying pattern. As a result, the updates become increasingly abstract and global as they move upward, with each layer embedding the full history of aggregation beneath it. A simplified two-layer form of this nested aggregation has been a foundational component in standard hierarchical FL~\cite{letaief, ourotherjournal, ourjournal, ourconf, hier_ota, purdue, chat, azimi3, lim, bennis, tony, energy_resource, feduc, place, wu, context, scheduling, wen, cost, poor, reen, prun2}.

	\textbf{Cloud Aggregation (Global Aggregation at Layer $N$):} At the final layer, the cloud server aggregates updates from all edge servers at layer $N-1$ to form a global model as
	\begin{align}
		\label{cloud_update}
		\mathbf{w}_{t+1} = \mathbf{w}_{t}+\sum_{i_{N-1} \in {\cal C}_{N-1}}^{} \frac{|{\cal C}_{N-1}^{i_{N-1}}|}{N_\text{tot}}Q_N(\mathbf{v}_{\tau_{N}}^{i_{N-1},t} - \mathbf{w}_{t}).
	\end{align}
	The coefficients $\tfrac{|{\cal C}_{N-1}^{i_{N-1}}|}{N_\text{tot}}$ mirror the global loss definition $F(\mathbf{w})$ in~\eqref{lossfunction_hier}. Consequently, by the recursive relation in~\eqref{edge_update}, this aggregation step approximates a global gradient descent update of the form $\mathbf{w}_{t+1} = \mathbf{w}_t - \mu \nabla F(\mathbf{w}_t)$, which corresponds to the standard optimization step for solving~\eqref{objective}. After aggregation, the updated global model is broadcasted back to all devices as
	\begin{align}
		\label{cloud_broadcast}
		\mathbf{w}_{0, \ldots,0}^{i,t+1} = \mathbf{w}_{t}, \forall i\in {\cal C}.
	\end{align}
	This procedure for $T$ global iterations is detailed in Algorithm 1.
	
	\begin{remark}
		An alternative way of expressing~\eqref{lossfunction_hier} is in terms of the
			average performance weighted by individual dataset sizes, i.e., \[
			F(\mathbf{w}) = 
			\frac{1}{\sum_{i = 1}^{N_\text{tot}}  D_{i}}
			\sum_{i_{N-1} \in \mathcal{C}_N} D_{N-1}^{i_{N-1}} F_{i_{N-1}}(\mathbf{w}),
			\]
			where
			\[
			F_{i_n}(\mathbf{w}) = 
			\frac{1}{ D_{n}^{i_{n}}}
			\sum_{i_{n-1} \in \mathcal{C}_n^{i_n}} D_{n-1}^{i_{n-1}} F_{i_{n-1}}(\mathbf{w}),
			\]
			and
			\[
			D_{n}^{i_{n}} = \sum_{i \in \mathcal{C}_1^{i_1}, \ldots, i_{n-1} \in \mathcal{C}_{n}^{i_{n}}} D_i,
			\]
			denotes the total number of samples under server $i_{n}$. Both formulations are valid~\cite{letaief, seyyed, hier_ota, purdue}: our original definition highlights equal treatment of devices, 
			giving the same priority to all nodes regardless of dataset size, due to the use of SGD with equal batch sizes. 
			In contrast, the alternative formulation assigns weights proportional to dataset size. 
			The algorithm and subsequent analysis can be directly extended to this weighted case 
			by replacing \[
			\frac{|{\cal C}_{n-1}^{i_{n-1}}|}{|{\cal C}_{n}^{i_n}|}
			\quad \text{with} \quad
			\frac{D_{n-1}^{i_{n-1}}}{D_{n}^{i_n}}, \;\; \forall n>1,
			\]
			and replacing $\mathbf{w}_{\tau,t_2,\ldots,t_{N}}^{i,t}$ with $D_i \mathbf{w}_{\tau,t_2,\ldots,t_{N}}^{i,t}$.
	\end{remark}
	\begin{remark}
		Adding more layers does not directly increase computational complexity, since each intermediate layer only performs aggregation and quantization without gradient computations. However, each layer $n$ introduces $\tau_n$ nested aggregation rounds, which trigger $\prod_{m=1}^{n}\tau_m$ local gradient computations at the device level. Hence, the overall computational cost is dominated by device-side updates and scales with $\mathcal{O}\left(\prod_{m=1}^{N}\tau_m\right)$.
	\end{remark}
	
	\vspace{-5pt}
	\subsection{Quantization}
	In the following, we characterize the quantization functions for different layers and present a practical example.
	
	\textbf{Assumption 1 (Unbiased Quantization):} The quantizers $Q_n, \forall n$ are unbiased and their variance grows proportionally to the square of the $l_2$-norm of their argument as
	\begin{eqnarray}
		\mathbb{E}\left\{Q_n(\mathbf{x})|\mathbf{x}\right\} = \mathbf{x},
	\end{eqnarray}
	\begin{eqnarray}
		\mathbb{E}\left\{\|Q_n(\mathbf{x})-\mathbf{x}\|^2|\mathbf{x}\right\} \leq q_n \|\mathbf{x}\|^2,
	\end{eqnarray}
	for any $\mathbf{x} \in \mathbb{R}^d$ and positive real constant $q_n$ as the variance of quantization error. This assumption ensures that quantization introduces zero-mean noise with bounded variance, which is a standard characteristic used for analytical tractability in communication-efficient FL, and is satisfied by widely used stochastic quantizers~\cite{alish,letaief,pedar,ourotherjournal}.
	
	Therefore, each model update---computed with respect to its previous state---is quantized and transmitted in~\eqref{edge_update} and~\eqref{cloud_update}, instead of transmitting the current model itself. This reduces the magnitude of the quantization input, thereby improving quantization quality.
	
	While the design or selection of specific quantization functions is beyond the scope of this work, our focus lies in analyzing the impact of the quantization variance associated with any unbiased quantizer on learning performance and optimization. For concreteness in the experiments presented in Section~VI, we adopt the following widely used example of unbiased quantizers.
	
	\textbf{Example for Quantizer~\cite{alish}.} For any variable $\mathbf{x}\in \mathbb{R}^d$, the
	quantizer $Q_n$: $\mathbb{R}^d \to \mathbb{R}^d$ is defined as below
	\begin{eqnarray}
		Q_n (\mathbf{x}) = \textrm{sign}\left\{\mathbf{x}\right\}\cdot\|\mathbf{x}\|\cdot \zeta(\mathbf{x},s_n),
	\end{eqnarray}
	where the $i$-th element of $\zeta(\mathbf{x},s_n)$, i.e., $\zeta_i(\mathbf{x}, s_n)$, is a random variable as
	\begin{align}
		\zeta_i(\mathbf{x},s_n) &= 
		\frac{l}{s_n}\ \textrm{with probability}\ 1-q\left({\frac{|x_i|}{\|\mathbf{x}\|}},s_n\right)\ \nonumber\\&\textrm{and} \
		\frac{l+1}{s_n}\ \textrm{with probability}\ q\left(\frac{|x_i|}{\|\mathbf{x}\|},s_n\right),
	\end{align}
	where $x_i$ is the $i$-th element of $\mathbf{x}$ and $q(a, s) = as-l$ for any $a\in[0, 1]$. In above, the tuning parameter $s_n$ corresponds to the number of quantization levels and $l \in [0, s_n)$ is an integer such that ${|x_i|}/{\|\mathbf{x}\|}\in [{l}/{s_n},{(l+1)}/{s_n}]$. As shown in~\cite{alish}, the variance \( q_n \) decreases as \( s_n \) increases.

	\begin{algorithm}[t]
		\small
		\caption{$N$-layer {\fontfamily{lmtt}\selectfont QMLHFL}}
		\begin{algorithmic}
			\State \textbf{Initialize} global model \( \mathbf{w}_0 \)
			\For{each global iteration \( t = 1, \dots, T \)}

			\For{each intra-layer iteration \( t_{N} = 1, \dots, \tau_{N} \)} \\\hspace{40pt}\vdots
			\State	\textbf{Nested Edge Aggregation at Layer \( n \):}
			\For{each intra-layer iteration \( t_{n+1} = 1, \dots, \tau_{n+1} \)}
			\For{each intra-layer iteration \( t_{n} = 1, \dots, \tau_{n} \)}
			\\\hspace{70pt}\vdots
			\State \textbf{Local Update at Devices:}
			\For{each intra-layer iteration \( t_1 = 1, \dots, \tau_1 \)}
			\State \hspace{0pt}Each device \( i \in {\cal C}\) updates its local model as~\eqref{local_update}
			\EndFor
			\\\hspace{70pt}\vdots
			\EndFor
			\State Each edge server \( i_n \in {\cal C}_n\) aggregates updates as~\eqref{edge_update}
			
			\State \textbf{Update propagation} as~\eqref{edge_broadcast}
			\EndFor\\\hspace{40pt}\vdots
			\EndFor
			\State \textbf{Cloud Aggregation} as in~\eqref{cloud_update}
			\State \textbf{Broadcast global model} as~\eqref{cloud_broadcast}
			\EndFor
		\end{algorithmic}
	\end{algorithm}
	
	\section{Convergence Analysis}
	The convergence rate of {\fontfamily{lmtt}\selectfont QMLHFL} is presented in the next theorem. The analysis is based on minimal standard assumptions that cover non-convex loss functions, which are commonly encountered in the literature (e.g.,~\cite{letaief,ourjournal,hier_ota,purdue,seyyed,conv2,conv3,conv4,prun2,castiglia,alish,pedar}), as outlined below.
	
	\textbf{Assumption 2 (Lipschitz-Continuous Gradient):} The gradient of the loss function $F(\mathbf{w})$ is characterized by Lipschitz continuity with a non-negative constant $L > 0$. This implies
	\begin{align}
		&F(\mathbf{w}_2) \leq F(\mathbf{w}_1) + \nabla F(\mathbf{w}_1)^T (\mathbf{w}_2-\mathbf{w}_1) + \frac{L}{2} \|\mathbf{w}_2 - \mathbf{w}_1\|^2,\\
		&\|\nabla F(\mathbf{w}_2)-\nabla F(\mathbf{w}_1)\| \leq L \|\mathbf{w}_2 - \mathbf{w}_1\|.
	\end{align}
	This assumption states that the gradient does not vary abruptly, reflecting the smooth behavior of standard, generally non-convex loss functions---such as those in neural networks---and enables bounding the change in the global loss after each local or aggregated update.
	
	\textbf{Assumption 3 (Gradient Variance Bound):} The local stochastic gradient $\nabla F(\mathbf{w},\boldsymbol \xi)$ serves as an unbiased estimator of the actual gradient $\nabla F(\mathbf{w})$, with its variance bounded as follows.
	\begin{align}
		\mathbb{E}\left\{\|\nabla F(\mathbf{w},\boldsymbol \xi) - \nabla F(\mathbf{w})\|^2\right\} \leq {\sigma^2}.
	\end{align}
	This assumption reflects the randomness introduced by mini-batch sampling at devices and is widely adopted in FL analyses to characterize the noise level in local gradient estimates.
	
	\begin{theorem}
		Assume that the learning rate $\mu$ satisfies the following condition:
		\begin{align}
			\label{convergence_condition}
			&1-{L^2\mu^2}\Biggl[\frac{\tau_1(\tau_1-1)}{2}+\sum_{n=2}^{N}\frac{\tau_n (\tau_n-1)}{2} \prod_{m=1}^{n-1}\tau_m^2 +q_1 \tau_2 \tau_1^2\nonumber
		\end{align}
		\begin{align}
			&+\sum_{n=1}^{N-2}\prod_{m=1}^{n+2}\tau_{m} \max_{}{\cal A}_{n}\Biggr]-{L} {\mu} \Biggl[\prod_{n=1}^{N}\tau_n +\frac{1}{N_\text{tot}} \max_{} {\cal A}_{N-1}\Biggr]\nonumber\\&\geq 0,
		\end{align}
		where ${\cal A}_n$ is a recursive function defined as
		\begin{align}
			{\cal A}_{n} &= |{\cal C}_{n}^{i_n}|\left(q_{n+1} \prod_{m=1}^{n+1}\tau_m + \frac{{|{\cal C}_{n-1}^{i_{n-1}}|}}{|{\cal C}_{n}^{i_n}|}(1+q_{n+1}){\cal A}_{n-1} \right), \nonumber\\&n \in \left\{2,\ldots, N-1\right\},
		\end{align}
		with
		\begin{align}
			{\cal A}_1 = {|{\cal C}_{1}^{i_1}|}\left(q_2 \tau_1 \tau_2+\frac{1}{|{\cal C}_{1}^{i_1}|} q_1 (1+q_2)\tau_1 \right).
		\end{align}
		The $\max$ operation is conducted with respect to the device counts. Then, the convergence rate of {\fontfamily{lmtt}\selectfont QMLHFL} after $T$ global iterations is bounded as
		\begin{align}
			\label{convergence_rate_asl}
			&\frac{1}{T}\sum_{t=0}^{T-1} \mathbb{E}\left\{\|\nabla F( {\mathbf{w}}_{t})\|^2\right\} \leq \frac{2(F(\mathbf{w}_0)-F(\mathbf{w}^*))}{\mu T\prod_{n=1}^{N}\tau_n}+\frac{L^2\mu^2}{2}\nonumber\\
			&\Biggl[ {(\tau_1-1)}+\sum_{n=1}^{N-1}\frac{C_{n}}{N_\text{tot}}{(\tau_{n+1}-1)}\prod_{m=1}^{n}(1+q_{m})\prod_{m=1}^{n}\tau_{m} \Biggr]\sigma^2\nonumber\\
			&+{L}\mu\frac{\sigma^2}{N_\text{tot}}\prod_{n=1}^{N}(1+q_{n}).
		\end{align}
	\end{theorem}
	\begin{IEEEproof}
		See Appendix.
	\end{IEEEproof}
	\begin{remark}
		The convergence rate bound in~\eqref{convergence_rate_asl} consists of two components: the first term characterizes the speed of convergence, while the remaining terms represent the post-convergence error. As the number of layers and intra-layer iteration counts $\tau_n, \forall n$ increase, the convergence speed improves; however, this also leads to a rise in the error term.
	\end{remark}
	\begin{remark}
		The convergence speed increases with $\prod_{n=1}^{N}\tau_n$, reflecting the effectiveness of the nested approach in leveraging the full hierarchical potential provided by multiple layers.
	\end{remark}
	\begin{remark}
		The condition in~\eqref{convergence_condition} defines the requirement on the learning rate $\mu$ necessary to guarantee convergence for {\fontfamily{lmtt}\selectfont QMLHFL}. This condition is influenced by the hierarchical architecture in its entirety, including the total number of devices and all device counts $|C_n^{i_n}|,\forall n$ associated with each edge server at every layer. However, the convergence rate itself depends only on two abstract characteristics of the architecture: the total number of devices and the number of edge servers per layer.
	\end{remark}
	\begin{remark}
		The condition in~\eqref{convergence_condition} involves several multiplicative terms composed of intra-layer iteration counts. These terms reveal that iteration counts in lower layers have a disproportionately larger effect on the convergence behavior, since they appear repeatedly and recursively in product terms involving deeper layers. As a result, increasing lower-layer \( \tau_n \) values can rapidly tighten the condition~\eqref{convergence_condition}. Therefore, any increase in intra-layer iteration counts must be accompanied by a corresponding decrease in the learning rate to ensure that the condition remains satisfied.
	\end{remark}
	\begin{remark}
		The recursive function ${\cal A}_n$ arises due to quantization and the nested aggregation structure inherent in {\fontfamily{lmtt}\selectfont QMLHFL}.
	\end{remark}
	\begin{remark}
		With a constant number of devices, an increased number of edge servers in any layer amplifies the error term in the convergence rate, resulting in diminished learning accuracy. While assigning devices or lower-layer edge servers to fewer upper-layer servers can improve performance at each global iteration, the overall hierarchical architecture, including the number of edge servers per layer, is often constrained by physical network limitations. Moreover, reducing the number of edge servers may increase the latency per iteration, as devices become farther from the servers, or servers from their corresponding higher-layer servers.
	\end{remark}
	\begin{remark}
		The bound in~\eqref{convergence_rate_asl} shows that quantization enters only through the factors
			$\prod_{m=1}^{n}(1+q_m)$ (inside the sum in the bracketed term) and $\prod_{n=1}^{N}(1+q_n)$ (the last term).
			Thus, as $N$ increases the quantization contribution is monotone nondecreasing. The sum term inside the brackets,
			$\sum_{n=1}^{N-1}\!\frac{C_n}{N_{\text{tot}}}(\tau_{n+1}-1)\prod_{m=1}^{n}(1+q_m)\prod_{m=1}^{n}\tau_m$,
			shows that lower-layer distortions (small $m$) propagate to all higher layers via the prefix products
			$\prod_{m=1}^{n}(1+q_m)$ and are further amplified by the iteration factors $\prod_{m=1}^{n}\tau_m$;
			higher-layer distortions affect fewer prefixes and thus accumulate less. In the small-quantization regime ($q_m \ll 1$), 
			$\prod_{m=1}^{n}(1+q_m) \approx 1+\sum_{m=1}^{n}q_m$ and 
			$\prod_{n=1}^{N}(1+q_n) \approx 1+\sum_{n=1}^{N}q_n$, 
			so the post-convergence error term in~\eqref{convergence_rate_asl} grows approximately linearly with $\sum_{n=1}^{N}q_n$. 
			If, instead, $q_n$ are not sufficiently small, the products increase exponentially with $N$ 
			(for example, if $q_n = q$ for all $n$, then $\prod_{n=1}^{N}(1+q_n) = (1+q)^N$), 
			which tightens the design constraints on quantization at deeper hierarchies.
			Practically, finer quantization (smaller $q_m$) should be used at lower layers---especially when $\tau_m$ are large---to control the cumulative factors and keep the depth-induced inflation moderate.
	\end{remark}

	We now highlight two noteworthy \textit{special cases}.
	
	a) When quantization is not a factor, and model parameters at each device or edge server in any layer are transmitted perfectly, i.e., $q_n = 0, \forall n=1,\ldots,N$, the convergence rate is simplified as follows.
	\begin{corollary}
		Assume no quantization is employed, and the learning rate $\mu$ satisfies the following condition:
		\begin{align}
			&1-{L^2\mu^2}\Biggl[\frac{\tau_1(\tau_1-1)}{2}+\sum_{n=2}^{N}\frac{\tau_n (\tau_n-1)}{2} \prod_{m=1}^{n-1}\tau_m^2\Biggr]\nonumber\\&-{L} {\mu} \prod_{n=1}^{N}\tau_n\geq 0.
		\end{align}
		Then, the convergence rate of {\fontfamily{lmtt}\selectfont QMLHFL} is bounded as
		\begin{align}
			&\frac{1}{T}\sum_{t=0}^{T-1} \mathbb{E}\left\{\|\nabla F( {\mathbf{w}}_{t})\|^2\right\} \leq \frac{2(F(\mathbf{w}_0)-F(\mathbf{w}^*))}{\mu T\prod_{n=1}^{N}\tau_n}+\frac{L^2\mu^2}{2}\nonumber\\&\Biggl[ {(\tau_1-1)}+\sum_{n=1}^{N-1}\frac{C_n}{N_\text{tot}}(\tau_{n+1} - 1) \prod_{m=1}^{n}\tau_m\Biggr]\sigma^2+{L}\mu\frac{1}{N_\text{tot}}\sigma^2.  
		\end{align}	
	\end{corollary}
	Although the non-quantized case of {\fontfamily{lmtt}\selectfont QMLHFL} is idealized, it has not been previously investigated. Therefore, the findings in this setting hold foundational value in research, aligning with prevailing trends in the field that consider perfect, non-quantized transmissions as a theoretical baseline~\cite{advanceFL}.

	b) When considering the case with two aggregation layers, denoted by \( N = 2 \), which corresponds to the hierarchical FL configuration found in the literature, the specific convergence rate from Theorem~1 is presented next.
	\begin{corollary}
		Consider a two-layer hierarchical FL, and assume that the learning rate $\mu$ satisfies the following condition:
		\begin{align}
			&1-{L^2\mu^2}\Biggl[\frac{\tau_1(\tau_1-1)}{2}+\tau_1^2 \frac{\tau_2(\tau_2-1)}{2}+q_1 \tau_2 \tau_1^2\biggr]-{L} {\mu} \Biggl[\tau_2 \tau_1\nonumber\\
			&+\frac{1}{N_\text{tot}} \max_{}\Biggl\{ |{\cal C}_{1}^{i_1}|\left(q_{2} \tau_1\tau_{2} + \frac{{1}}{|{\cal C}_{1}^{i_1}|}(1+q_2)q_{1} \tau_1\right)\Biggr\}\Biggr]\geq 0,
		\end{align}
		Then, the convergence rate is bounded as
		\begin{align}
			\frac{1}{T}\sum_{t=0}^{T-1}& \mathbb{E}\left\{\|\nabla F( {\mathbf{w}}_{t})\|^2\right\} \leq \frac{2(F(\mathbf{w}_0)-F(\mathbf{w}^*))}{\mu T\tau_2 \tau_1}\nonumber\\
			&+\frac{L^2\mu^2}{2}\Biggl[ {(\tau_1-1)}+ (1+q_1) {(\tau_2-1)} \tau_1  \frac{C_1}{N_\text{tot}}\Biggr]\sigma^2\nonumber\\
			&+{L}\mu\frac{1}{N_\text{tot}}(1+q_{2}) (1+q_1)\sigma^2.  
		\end{align}
	\end{corollary}
	While such convergence rate bound reduces to that presented in~\cite[Eq.~5]{letaief}, the convergence conditions are structurally different, with our condition being tighter. This is because, although both conditions depend on the total number of devices, our condition explicitly incorporates device counts as architectural parameters and captures their impact. In contrast, the condition in~\cite[Eq.~4]{letaief} does not exhibit any architectural dependency. This difference stems from the tighter bounds derived through our analytical approach, as detailed in Appendix. This underscores the novelty of our convergence analysis, even for the specialized two-layer case, which is notably simpler than the general $N$-layer scenario.
	\vspace{-5pt}
	\section{Learning Optimization}
	The intra-layer iteration counts \( \tau_n \), for all \( n = 1, \ldots, N \), which are fundamental parameters in {\fontfamily{lmtt}\selectfont QMLHFL}, can be selected to minimize the convergence rate bound in~\eqref{convergence_rate_asl}, thereby improving the overall learning performance. Convergence rate bounds have also been employed for optimization in prior FL studies (e.g.,~\cite{conv2, conv3, conv4}). However, obtaining accurate values for \( F(\mathbf{w}^*) \), \( L \), and \( \sigma^2 \) in~\eqref{convergence_rate_asl} generally requires prior statistical knowledge of the local learning models---information that may not be readily available in many practical scenarios. To overcome this limitation and ensure universal applicability, we propose a weighted objective function that combines the first term in~\eqref{convergence_rate_asl} (representing convergence speed) and the second term (associated with the post-convergence error), moderated by a scaling parameter \( 0 \leq \alpha \leq 1 \). Both terms are normalized with respect to their reference values to ensure unit consistency, although for simplicity, the normalization factors are omitted in the subsequent expressions. The parameter \( \alpha \) is treated as a hyperparameter, allowing the trade-off between faster convergence and lower post-convergence error to be tuned based on application requirements.
	
	Meanwhile, the run-time required to complete $T$ global iterations of {\fontfamily{lmtt}\selectfont QMLHFL} is given by $t_\text{d}T$, where the latency per global iteration under synchronous transmission\footnote{Synchronicity is widely adopted in the FL literature~\cite{mcmahan, letaief, alish, pedar, ourotherjournal, ourjournal, ourconf, hier_ota, purdue, conv2, conv3, conv4, prun2, chat, lim, bennis, tony, energy_resource, feduc, scheduling, place, wu, context, cost, castiglia, muulti, azimi3, reen, poor, wen, advanceFL,seyyed}.} is 
	\begin{align}
		\label{latency}
		t_\text{d} = &\prod_{n=1}^{N}\tau_n t_{\text{CP}}+\prod_{n=2}^{N}\tau_n t_\text{DE}+\sum_{n=2}^{N-1}\prod_{m=n+1}^{N}\tau_m t_{\text{E}_{n-1,n}}+\nonumber\\& t_{\text{E}_{N-1,N}}.
	\end{align}
	In~\eqref{latency}, $t_\text{CP}$ represents the maximum computation time among all devices, corresponding to the slowest device, while $t_\text{DE}$ and $t_{\text{E}_{n-1,n}}$ denote the communication times between each device and its respective edge server in layer 1 and between each edge server in layer $n-1$ and its respective edge server in layer $n$, respectively. From~\cite{letaief,bennis,conv3,conv4,tony, scheduling,cost}, these parameters can be obtained as
	$
		t_\text{CP} = {cb}/{f_\text{min}},
		$
	$
	t_\text{DE} = {d_\text{b}}/{\bigl(W\log_2\bigl(1+\frac{ph}{N_0}\bigr)\bigr)},
	$
	where $c$ is the number of CPU cycles to
	execute one sample of data, $f_\text{min} = \min_i f_i$, where $f_i$ is the CPU cycle frequency of device $i$, denotes the minimum CPU frequency across all devices, $d_\text{b}$ is the model size in bits, $W$ is the channel bandwidth, $h$ is the channel gain, $p$ is the transmission power, and $N_0$ is the noise power. 
	
	Thus, ensuring task completion within a predefined deadline $T_{\text{d}}$ is essential.\footnote{We assume the hardware-related parameters \( c \) and \( f_i \) for each device \( i \), as well as the deadline \( T_\text{d} \), are fixed and known, in line with~\cite{letaief, bennis, tony, scheduling,cost,conv3,conv4}.} This requirement leads to the following optimization problem formulation.
	\begin{align}
		\label{main_opt}
		&\min_{\tau_1,\cdots,\tau_N}\alpha \prod_{n=1}^{N}{\tau_n^{-1}} + (1-\alpha)\Biggl[ {(\tau_1-1)}+\nonumber\\
		&\sum_{n=1}^{N-1}\frac{C_{n}}{N_\text{tot}}{(\tau_{n+1}-1)}\prod_{m=1}^{n}(1+q_{m})\prod_{m=1}^{n}\tau_{m} \Biggr],
	\end{align}
	subject to
	\begin{align}
		&\prod_{n=1}^{N}\tau_n t_{\text{CP}}+\prod_{n=2}^{N}\tau_n t_\text{DE}+\sum_{n=2}^{N-1}\prod_{m=n+1}^{N}\tau_m t_{\text{E}_{n-1,n}}+ t_{\text{E}_{N-1,N}} \nonumber\\
		&\leq \frac{T_{\text{d}}}{T}. \nonumber
	\end{align}
	This is a nonlinear optimization problem, which can be reformulated as 
	\begin{align}
		\label{GP}
		&\min_{\boldsymbol \tau} J^+ (\boldsymbol \tau) - J^- (\boldsymbol \tau),
	\end{align}
	subject to $G(\boldsymbol{\tau}) \leq 1$,
	where $\boldsymbol{\tau} = [\tau_1,\ldots,\tau_N]^\top$, and
	\begin{align}
		&J^+(\boldsymbol{\tau}) = \alpha \prod_{n=1}^{N}{\tau_n^{-1}} + (1-\alpha)\nonumber\\&\Biggl[\tau_1+\sum_{n=1}^{N-1}\frac{C_{n}}{N_\text{tot}}\prod_{m=1}^{n}(1+q_{m})\prod_{m=1}^{n+1}\tau_{m}\Biggr],
	\end{align}
	\begin{align}
		\label{Jminus}
		&J^-(\boldsymbol{\tau}) = (1-\alpha)\Biggl[1+\sum_{n=1}^{N-1}\frac{C_{n}}{N_\text{tot}}\prod_{m=1}^{n}(1+q_{m})\prod_{m=1}^{n}\tau_{m}\Biggr],
	\end{align}
	\begin{align}
		\label{Gconstraint}
		G(\boldsymbol{\tau}) &= \frac{T}{T_\text{d}}\biggl(\prod_{n=1}^{N}\tau_n t_{\text{CP}}+\prod_{n=2}^{N}\tau_n t_\text{DE}+\sum_{n=2}^{N-1}\prod_{m=n+1}^{N}\tau_m t_{\text{E}_{n-1,n}}\nonumber\\&+ t_{\text{E}_{N-1,N}}\biggr).
	\end{align}
	Problem~\eqref{GP} is a complementary
	geometric programming~\cite{cgp}. One approach to solve~\eqref{GP} is to convert it into a
	sequence of standard geometric programming problems that can be efficiently solved
	to achieve a global solution~\cite{chiang}. This
	means that the variables in the optimization problem should
	be all positive, and the objective function and the constraints
	should be posynomials. For this, using an auxiliary
	variable $\delta$, we reformulate~\eqref{GP} as
	\begin{align}
		\label{posy}
		&\min_{\boldsymbol \tau, \delta} \delta,
	\end{align}
	subject to $G(\boldsymbol{\tau}) \leq 1$ and
	\begin{align}
		&J(\boldsymbol\tau, \delta) = \frac{J^+ (\boldsymbol \tau)}{ J^- (\boldsymbol \tau)+\delta} \leq 1.\nonumber
	\end{align}
	In~\eqref{posy}, $G(\boldsymbol{\tau})$ is posynomial, whereas $J(\boldsymbol\tau, \delta)$ is non-posynomial. The arithmetic-geometric mean approximation (AGMA) can
	be applied to transform $J(\boldsymbol\tau, \delta)$ to
	a posynomial function~\cite{chiang}. The approximated function is given by
	\begin{align}
		\label{Jtilde}
		\tilde J(\boldsymbol\tau, \delta) = \frac{J^+ (\boldsymbol \tau)}{\tilde J^- (\boldsymbol \tau,\delta)},
	\end{align}
	where
	\begin{align}
		&\tilde J^- (\boldsymbol \tau,\delta) = \left(\frac{1-\alpha+\delta}{\beta_0}\right)^{\beta_0}\nonumber\\&\prod_{k=1}^{N-1} \left(\frac{(1-\alpha)\frac{C_{k}}{N_\text{tot}}\prod_{m=1}^{k}(1+q_{m})\prod_{m=1}^{k}\tau_{m} }{\beta_k}\right)^{\beta_k}.
	\end{align}
	Thus, we can convert~\eqref{posy} into the subsequent problems, where in each iteration $l$, the initial step involves solving the following geometric programming problem.
	\begin{align}
		\label{posy2}
		&\min_{\boldsymbol \tau^l, \delta^l} \delta^l,
	\end{align}
	subject to $G(\boldsymbol{\tau}^l) \leq 1$ and
	\begin{align}
		&\tilde J(\boldsymbol\tau^l, \delta^l) \leq 1,\nonumber
	\end{align}
	and in the second step, the following update is considered.
	\begin{align}
		\label{beta0}
		\beta_0^{l+1} = \frac{1-\alpha+\delta^l}{J^- (\boldsymbol \tau^l)+\delta^l},
	\end{align}
	\begin{align}
		\label{betak}
		\beta_k^{l+1} = \frac{(1-\alpha)\frac{C_{k}}{N_\text{tot}}\prod_{m=1}^{k}(1+q_{m})\prod_{m=1}^{k}\tau_{m}^l}{J^- (\boldsymbol \tau^l)+\delta^l}, \forall k \geq 1.
	\end{align}
	
	\begin{algorithm}[t]
		\caption{Intra-Layer Iteration Count Optimization}
		\label{alg:tau_opt}
		\begin{algorithmic}[0]
			\State \textbf{Input:} Hyperparameter $\alpha$, time parameters $t_\text{CP}$, $t_\text{DE}$, $t_{\text{E}_{n-1,n}}$, deadline $T_d$, maximum iteration count $L_{\max}$, tolerance $\epsilon$
			\State \textbf{Initialize:} Set $\boldsymbol{\tau}^{0} \succ \mathbf{0}$, $\delta^{0} > 0$, $l \gets 0$
			\Repeat
			\State Compute $J^{-}(\boldsymbol{\tau}^{l})$ as in~\eqref{Jminus}
			\State Update coefficients $\beta_0^{l+1}$ and $\beta_k^{l+1}$ as in~\eqref{beta0} and~\eqref{betak}
			\State Compute $\tilde J (\boldsymbol \tau^l,\delta^l)$ and $G(\boldsymbol\tau^l)$ as in~\eqref{Jtilde} and~\eqref{Gconstraint}
			\State Solve the geometric program~\eqref{posy2} to obtain $(\boldsymbol{\tau}^{l+1},\delta^{l+1})$
			\State $l \gets l + 1$
			\Until{convergence: $|\delta^{l} - \delta^{l-1}| < \epsilon$ or $l = L_{\max}$}
		\end{algorithmic}
	\end{algorithm}
	
	This optimization procedure is detailed in Algorithm 2. As a \textit{special case}, in computation-limited systems where communication times are negligible compared to computation time, i.e., $t_\text{CP} \gg t_\text{DE},\ t_{\text{E}_{n-1,n}},\forall n,$ the optimization problem in~\eqref{main_opt} can be equivalently reformulated as follows:
	\begin{align}
		\label{alt_opt}
		&\min_{\tau_1,\cdots,\tau_N} {(\tau_1-1)}+\sum_{n=1}^{N-1}\frac{C_{n}}{N_\text{tot}}{(\tau_{n+1}-1)}\prod_{m=1}^{n}(1+q_{m})\prod_{m=1}^{n}\tau_{m},
	\end{align}
	subject to 
	\begin{align}
		\prod_{n=1}^{N}\tau_n = \frac{T_{\text{d}}}{Tt_{\text{CP}}}.\nonumber
	\end{align} 
	It can be readily shown that the optimal solution to~\eqref{alt_opt} is
	\begin{align}
		\tau_l^* = \frac{T_{\text{d}}}{Tt_{\text{CP}}},\ \tau_n^* = 1, \forall n \neq l,
	\end{align}
	where
	\begin{align}
		\label{selection}
		l = \arg \min_n \frac{C_{n}}{N_\text{tot}}\prod_{m=1}^{n}(1+q_{m}).
	\end{align}
	Thus, when model parameters are transmitted without quantization, i.e., $q_n = 0, \forall n$, the highest layer is always selected as the solution to~\eqref{selection}. Otherwise, the selection depends on the number of edge servers at each layer and the product of quantization error variances across preceding layers.
	
	\vspace{-5pt}

	\vspace{0pt}
	\section{Experimental Results}
	
	We evaluate the performance of {\fontfamily{lmtt}\selectfont QMLHFL} using two benchmark datasets, MNIST and CIFAR-10. The experiments are performed under 3-layer, 4-layer, and 6-layer hierarchical setups, while 1-layer ({\fontfamily{lmtt}\selectfont FedAvg}~\cite{mcmahan, pedar}) and 2-layer ({\fontfamily{lmtt}\selectfont Hier-Local-QSGD}~\cite{letaief}) setups are included as baselines representing traditional flat and hierarchical FL, respectively.\footnote{Due to the limited computational power of our simulation system and the theoretical nature of this work, we were unable to provide experiments with setups involving more layers.} It is assumed that each device $i$ has a distinct CPU cycle frequency $f_i$ that is uniformly and randomly drawn from the range $[0.5, 2]~\text{GHz}$. The learning rate is set to $\mu = 0.01$. The performance is measured in terms of test accuracy, i.e., the classification accuracy on a held-out test set, and training loss at each global iteration
	$t$.
	
	\subsection{Data Distribution and Hierarchical Setups}
	To simulate real-world data heterogeneity, the dataset is partitioned across devices such that each device receives a random number of local training samples uniformly drawn from the range [500, 1500]. The mini-batch size is set to $b = 40$. The data allocation follows three scenarios: samples drawn from 2 randomly selected classes (Case 1), from 6 randomly selected classes (Case 2), or uniformly from all 10 classes (Case 3).
	
	We consider a hierarchical architecture with 96 devices at the device layer ($\mathcal{C}$), 32 edge servers at the first edge layer ($\mathcal{C}_1$), 16 edge servers at the second edge layer ($\mathcal{C}_2$), 8 edge servers at the third edge layer ($\mathcal{C}_3$), 4 edge servers at the fourth edge layer ($\mathcal{C}_4$), 2 edge servers at the fifth edge layer ($\mathcal{C}_5$), and a cloud server. 
		The number of quantization levels at these layers are set as $s_1 = 4$, $s_2 = 6$, $s_3 = 8$, $s_4 = 10$, $s_5 = 12$, and $s_6 = 14$, respectively.\footnote{In our work, the quantization error variance $q$ is measured numerically.} 
		In the \textbf{6-layer setup}, each group of 3 devices is connected to one server in $\mathcal{C}_1$, and each higher-layer server is connected to two servers from the layer below, continuing this two-to-one connection pattern up to the cloud.
	
	For the reduced-depth setups derived from this baseline:
		\begin{itemize}
			\item \textbf{4-layer setup:} removing $\mathcal{C}_1$ and $\mathcal{C}_2$, each group of 12 devices ($3 \times 2 \times 2$) is connected to one server in $\mathcal{C}_3$, followed by the two-to-one connections up to the cloud.
			\item \textbf{3-layer setup:} removing $\mathcal{C}_1$---$\mathcal{C}_3$, each group of 24 devices ($3 \times 2 \times 2 \times 2$) is connected to one server in $\mathcal{C}_4$, followed by the two-to-one connections.
			\item \textbf{2-layer setup:} removing $\mathcal{C}_1$---$\mathcal{C}_4$, each $\mathcal{C}_5$ server is connected to 48 devices ($3 \times 2 \times 2 \times 2 \times 2$) and then to the cloud.
			\item \textbf{1-layer setup:} removing $\mathcal{C}_1$---$\mathcal{C}_5$, all 96 devices ($3 \times 2 \times 2 \times 2 \times 2 \times 2$) are directly connected to the cloud server.
		\end{itemize}
		Removing lower edge layers increases the characteristic device-server distance (and thus weakens the effective channel) relative to the baseline 6-layer deployment. 
		We denote the corresponding multiplicative distance factors by $\kappa_4$ (removing $\mathcal{C}_1$---$\mathcal{C}_2$), $\kappa_3$ (removing $\mathcal{C}_1$---$\mathcal{C}_3$), $\kappa_2$ (removing $\mathcal{C}_1$---$\mathcal{C}_4$), and $\kappa_1$ (removing $\mathcal{C}_1$---$\mathcal{C}_5$), all measured with respect to the reference channel gain $h$ reported in Table~II.
	
	\begin{figure*}
		\centering
		\begin{subfigure}[t]{0.32\linewidth}
			\centering
			\includegraphics[width=\linewidth]{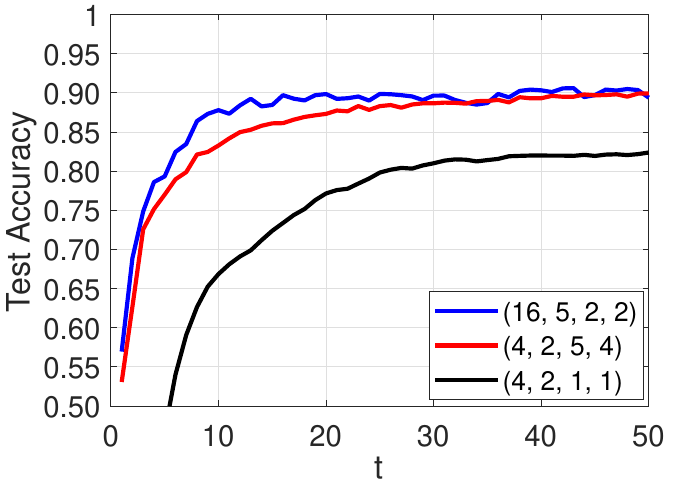}
			\caption*{(a)}
		\end{subfigure}
		\hfill
		\begin{subfigure}[t]{0.32\linewidth}
			\centering
			\includegraphics[width=\linewidth]{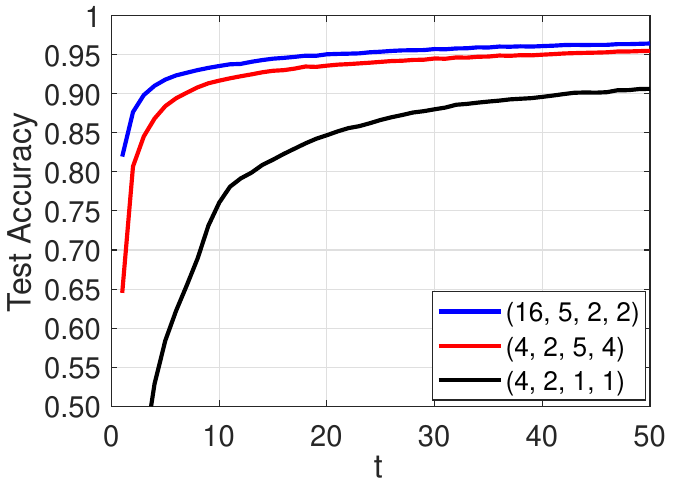}
			\caption*{(b)}
		\end{subfigure}
		\hfill
		\begin{subfigure}[t]{0.32\linewidth}
			\centering
			\includegraphics[width=\linewidth]{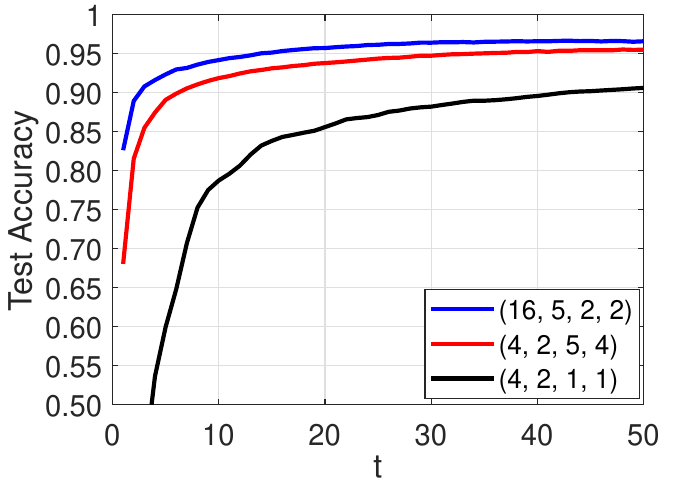}
			\caption*{(c)}
		\end{subfigure}
		
		\vspace{0.5em}
		\caption{\small Test accuracy of the 4-layer {\fontfamily{lmtt}\selectfont QMLHFL} under the data heterogeneity cases: (a) Case 1, (b) Case 2, and (c) Case 3.}
	\end{figure*}
	
	\begin{figure*}
		\centering
		\begin{subfigure}[t]{0.32\linewidth}
			\centering
			\includegraphics[width=\linewidth]{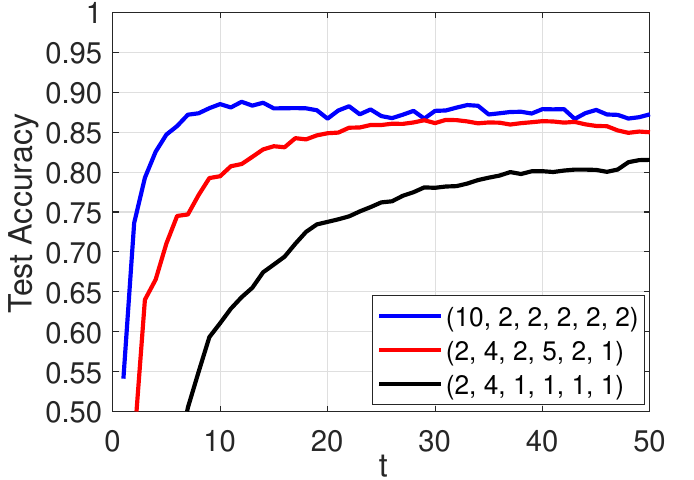}
			\caption*{(a)}
		\end{subfigure}
		\hfill
		\begin{subfigure}[t]{0.32\linewidth}
			\centering
			\includegraphics[width=\linewidth]{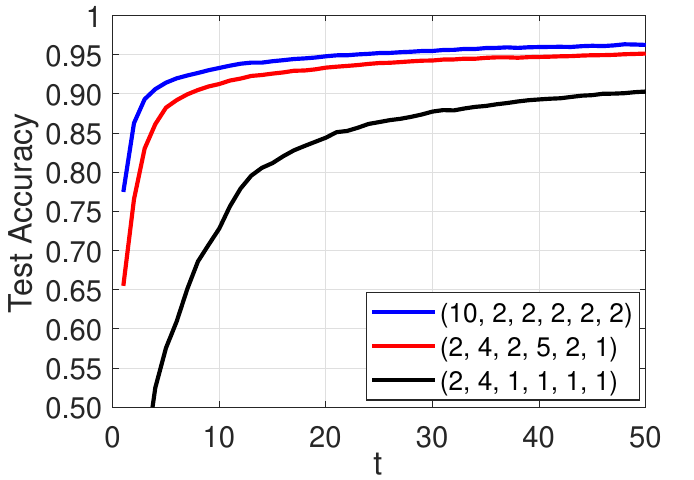}
			\caption*{(b)}
		\end{subfigure}
		\hfill
		\begin{subfigure}[t]{0.32\linewidth}
			\centering
			\includegraphics[width=\linewidth]{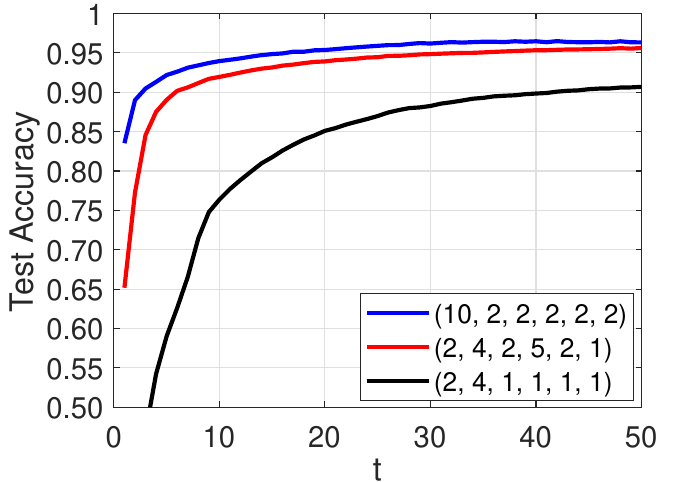}
			\caption*{(c)}
		\end{subfigure}
		
		\vspace{0.5em}
		\caption{\small Test accuracy of the 6-layer {\fontfamily{lmtt}\selectfont QMLHFL} under the data heterogeneity cases: (a) Case 1, (b) Case 2, and (c) Case 3.}
	\end{figure*}
	
	\begin{figure*}
		\centering
		\begin{subfigure}[t]{0.32\linewidth}
			\centering
			\includegraphics[width=\linewidth]{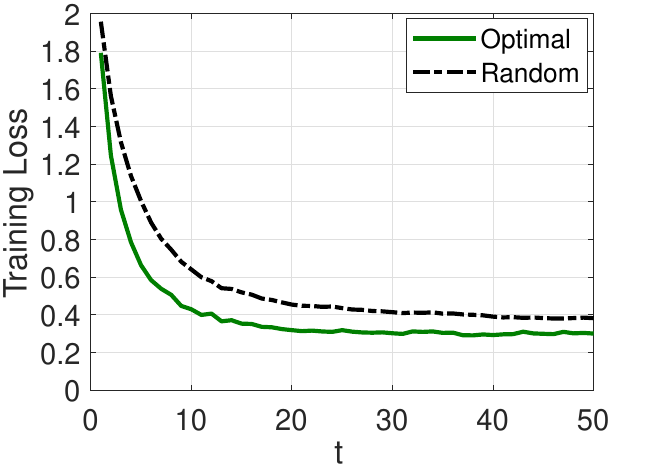}
			\caption*{(a)}
		\end{subfigure}
		\hfill
		\begin{subfigure}[t]{0.32\linewidth}
			\centering
			\includegraphics[width=\linewidth]{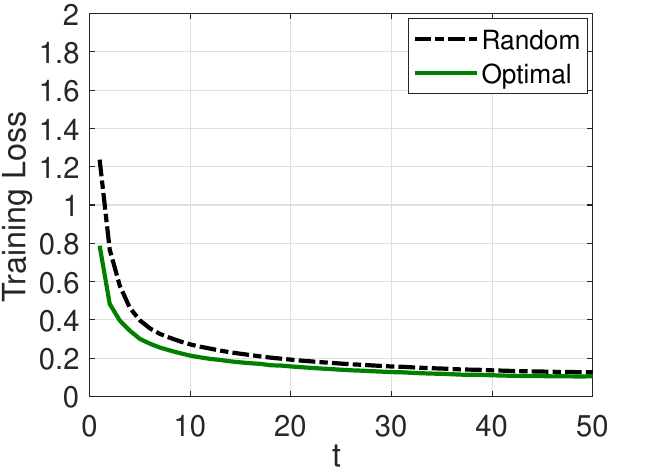}
			\caption*{(b)}
		\end{subfigure}
		\hfill
		\begin{subfigure}[t]{0.32\linewidth}
			\centering
			\includegraphics[width=\linewidth]{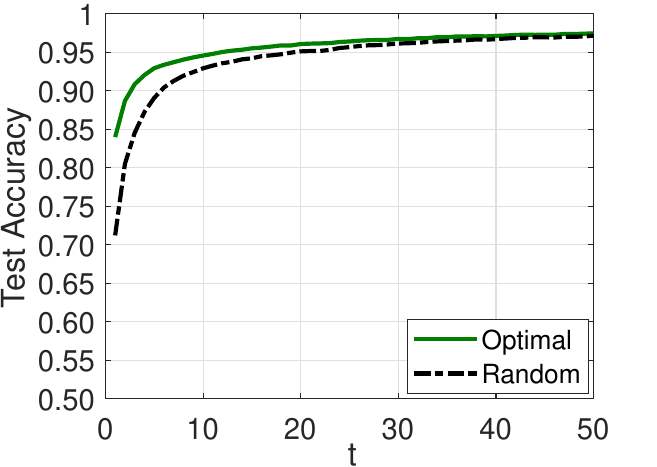}
			\caption*{(c)}
		\end{subfigure}
		
		\vspace{0.5em}
		\caption{\small Performance of the 6-layer {\fontfamily{lmtt}\selectfont QMLHFL} under the data heterogeneity cases: (a) training loss for Case 1, (b) training loss for Case 2, and (c) test accuracy for Case 3.}
	\end{figure*}

	\begin{figure*}[t]
		\centering
		\begin{subfigure}[t]{0.36\linewidth}
			\centering
			\includegraphics[width=\linewidth]{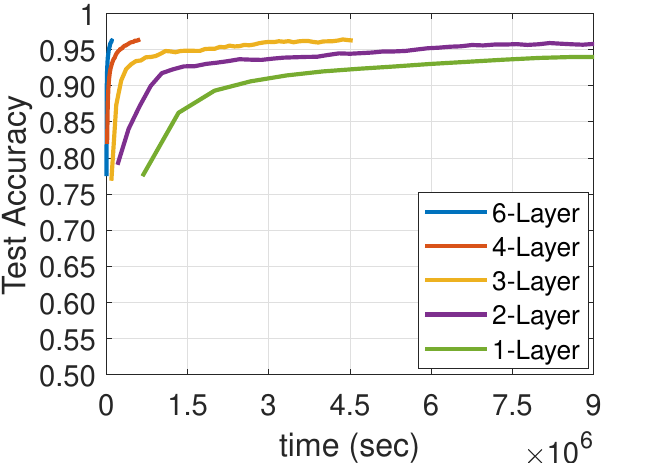}
			\caption*{(a)}
		\end{subfigure}
		\hfill
		\begin{subfigure}[t]{0.36\linewidth}
			\centering
			\includegraphics[width=\linewidth]{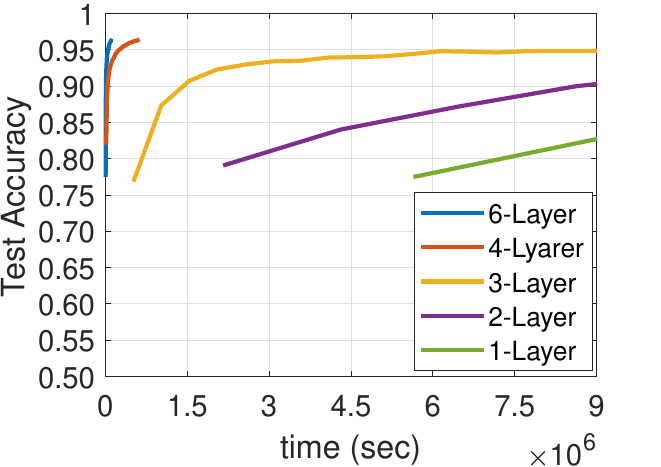}
			\caption*{(b)}
		\end{subfigure}
		
		\vspace{0.5em}
		\caption{\small Test accuracy as a function of run-time: 
				(a) ($\kappa_1 = 80$, $\kappa_2 = 25$, $\kappa_3 = 15$, $\kappa_4 = 10$) and (b) ($\kappa_1 = 150$, $\kappa_2 = 50$, $\kappa_3 = 25$, $\kappa_4 = 10$).}
	\end{figure*}
	
	\subsection{MNIST Evaluation with MLP}
	We begin with the MNIST dataset, employing a multi-layer perceptron (MLP) model comprising an input layer of 784 neurons, two hidden layers with 128 and 64 ReLU-activated neurons (each followed by a 30\% dropout), and a 10-neuron softmax output layer. The results are visualized in the following figures:
	
	Figs.~2(a)--(c) present test accuracy results for 4-layer {\fontfamily{lmtt}\selectfont QMLHFL} with three $(\tau_1, \tau_2, \tau_3, \tau_4)$ configurations, evaluated under the three data heterogeneity cases. Similarly, Figs.~3(a)--(c) show test accuracy results for 6-layer {\fontfamily{lmtt}\selectfont QMLHFL} with three $(\tau_1, \tau_2, \tau_3, \tau_4, \tau_5, \tau_6)$ configurations, under the same cases. As observed, both with varying numbers of layers have successfully converged and demonstrated satisfactory learning accuracy, even under conditions of high data heterogeneity. Furthermore, increasing the product $\tau_1\tau_2\tau_3\tau_4$ in the 4-layer setup and $\tau_1\tau_2\tau_3\tau_4\tau_5\tau_6$ in the 6-layer setup leads to faster convergence, in line with \textit{Remark 4}. Notably, increasing the number of intra-layer iterations in the higher layers ($\tau_n$ for $n>2$, representing deeper hierarchy) proves highly effective, leading to substantial performance improvement and underscoring the pivotal roles of these deeper layers in hierarchical learning. When the product is the same for both setups, they exhibit nearly identical convergence speeds. However, under high data heterogeneity, the 4-layer setup---with more devices assigned to each first-layer edge server---achieves faster convergence and higher test accuracy. This is attributed to each edge server training on a larger, more diverse dataset, which helps mitigate the effects of heterogeneity and improves overall learning performance.
	
	Figs. 4(a)--(c) present the training loss and test accuracy performance of the 6-layer {\fontfamily{lmtt}\selectfont QMLHFL} using both optimized and randomly selected feasible $(\tau_1, \tau_2, \tau_3, \tau_4, \tau_5, \tau_6)$ tuples, each meeting a deadline constraint of $T_\text{d} = 400T\ \text{sec}$. Training loss is reported for Case 1 and Case 2 data distributions, while test accuracy is shown for Case 3, ensuring a diverse and comprehensive evaluation. The optimization follows the method described in Section V with $\alpha = 0.6$, and the corresponding run-time parameters are listed in Table II. As illustrated, the optimized configuration consistently achieves faster convergence, higher test accuracy, and lower training loss compared to the random selection, thereby demonstrating the effectiveness of the proposed learning optimization framework.
	
	In Figs.~5(a)--(b), the test accuracy of the 3-layer, 4-layer, and 6-layer {\fontfamily{lmtt}\selectfont QMLHFL} is compared with traditional 2-layer hierarchical and 1-layer (non-hierarchical) FL under the Case~2 data distribution, for two different sets of $(\kappa_1, \kappa_2, \kappa_3, \kappa_4)$. Specifically, the 6-layer setup is configured with $(\tau_1 = 10, \tau_2 = 2, \tau_3 = 2, \tau_4 = 2, \tau_5 = 2, \tau_6 = 2)$, the 4-layer setup with $(\tau_1 = 16, \tau_2 = 5, \tau_3 = 2, \tau_4 = 2)$, the 3-layer setup with $(\tau_1 = 8, \tau_2 = 10, \tau_3 = 4)$, the 2-layer setup with $(\tau_1 = 20, \tau_2 = 16)$, and the 1-layer setup with $(\tau_1 = 320)$, all maintaining the same product. Each curve is plotted until convergence, while those continuing to the end of the timing range indicate that convergence was not yet achieved. Considering a path-loss exponent of 3.4, the effective channel gains in the $n$-layer setup are $(\kappa_n)^{-3.4} h$, resulting in distinct values of $t_\text{DE}$. Following a similar per-iteration latency evaluation as in~\eqref{latency}, and accounting for the removed layers in each setup, the run-time can be expressed as $t t_{\text{d}}$, where $t_{\text{d}}$ represents the latency calculated for each setup based on the parameters in Table~II. The communication times between edge servers are assumed identical across all setups and are evaluated using $t_\text{DE}$ from the 6-layer setup, as specified in the table. As observed, increasing the number of layers significantly enhances both convergence speed and accuracy at any given run-time before convergence, demonstrating the clear superiority of {\fontfamily{lmtt}\selectfont QMLHFL} over traditional 2-layer and 1-layer FL. Moreover, even a slight increase in the distance ratio $\kappa_n$ in any $n$-layer setup leads to a substantial widening of this performance gap. For example, in the case with $(\kappa_1 = 80, \kappa_2 = 25, \kappa_3 = 15, \kappa_4 = 10)$, the total run-time in the 1-layer, 2-layer, and 3-layer setups becomes approximately 326, 100, and 45 times that of the 6-layer setup, respectively. In another case with $(\kappa_1 = 150, \kappa_2 = 50, \kappa_3 = 25, \kappa_4 = 10)$, the corresponding ratios further increase to about 2760, 1054, and 250 times the run-time of the 6-layer setup.
	
	While the same behavior as in Figs.~5(a)--(b) is observed for the other data distributions in Cases~1 and~3, a more careful investigation is required after convergence. The measured post-convergence test accuracies for the 1-layer to 6-layer setups are summarized in Table~III under the three data heterogeneity cases. By setting $L = 5$ and $\sigma^2 = 10^{-6}$ as in~\cite{amiri}, the post-convergence error terms in the convergence rate expression~\eqref{convergence_rate_asl} are calculated as $0.2566$, $1.079\times10^{-4}$, $3.217\times10^{-6}$, $1.141\times10^{-7}$, and $8.346\times10^{-9}$, respectively. These results not only confirm the accumulation-error behavior discussed in \textit{Remark~9} but also explain the slight improvement in post-convergence test accuracy as the number of layers decreases across all cases. Moreover, as data heterogeneity increases, the performance gap between setups with fewer layers and those with more layers becomes slightly larger, primarily because a higher number of devices are assigned to each first-layer edge server. Therefore, {\fontfamily{lmtt}\selectfont QMLHFL} is not only necessitated by the physical constraints of network configuration but also achieves substantially faster convergence, particularly since removing layers considerably increases the average distance between devices and their first-layer servers, while maintaining almost identical performance after convergence.
	
	\begin{table}[t]
		\centering
		\caption{Run-time Parameters}
		\vspace{-6pt}
		\resizebox{9cm}{!}{%
			\begin{tabular}{|l|l|l|l|l|l|l|l|l|l|}
				\hline
				{$W$} & {$p$} & {$N_0$} & {$c$} & {$h$} & {$t_{\text{E}_{1,2}}$} & {$t_{\text{E}_{2,3}}$} & {$t_{\text{E}_{3,4}}$} & $t_{\text{E}_{4,5}}$ & $t_{\text{E}_{5,6}}$ \\
				\hline
				$1\,\text{MHz}$ & $0.5\,\text{W}$ & $10^{-10}\,\text{W}$ & $0.25\times 10^{9}\,\ \text{cycles}$ & $10^{-8}$ & $10\,t_{\text{DE}}$ & $20\,t_{\text{DE}}$ & $30\,t_{\text{DE}}$ & $40\,t_{\text{DE}}$ & $50\,t_{\text{DE}}$ \\
				\hline
			\end{tabular}%
		}
		\vspace{-6pt}
	\end{table}
	
	\begin{table}[t]
		\centering
		\caption{Final MNIST test accuracy across $n$-layer FL}
		\begin{tabular}{|c|c|c|c|c|c|}
			\hline
			\textbf{Case} & $n=1$ & $n=2$ & $n = 3$ & $n = 4$ & $n = 6$ \\
			\hline
			1 & 0.9257 & 0.9167 & 0.9112 & 0.8940 & 0.8780 \\
			2 & 0.9712 & 0.9696 & 0.9683 & 0.9603 & 0.9586 \\
			3 & 0.9754 & 0.9741 & 0.9737 & 0.9689 & 0.9678 \\
			\hline
		\end{tabular}
	\end{table}

	\begin{table}[t]
		\centering
		\caption{Test accuracy on CIFAR-10 at $t = 100$}
		\begin{tabular}{|c|c|c|c|}
			\hline
			\textbf{Case} & $s_1 = 3$, $s_2 = 5$ & $s_1 = 6$, $s_2 = 8$ & $s_1 = 8$, $s_2 = 10$ \\
			\hline
			1 & 0.5124 & 0.6298 & 0.6645 \\
			2 & 0.7383 & 0.8176 & 0.8602 \\
			3 & 0.7445 & 0.8389 & 0.8747 \\
			\hline
		\end{tabular}
	\end{table}
	
	\begin{table}[t]
		\centering
		\caption{Test accuracy on CIFAR-10 at $t = 100$}
		\begin{tabular}{|c|c|c|c|}
			\hline
			\textbf{Case} & $s_5 = 3$, $s_6 = 5$ & $s_5 = 6$, $s_6 = 8$ & $s_5 = 8$, $s_6 = 10$ \\
			\hline
			1 & 0.5487 & 0.5645 & 0.5753 \\
			2 & 0.7507 & 0.7753 & 0.7820 \\
			3 & 0.7729 & 0.7881 & 0.8006 \\
			\hline
		\end{tabular}
	\end{table}
	\vspace{-5pt}
	\subsection{CIFAR-10 Evaluation with Deep CNN}
	To evaluate {\fontfamily{lmtt}\selectfont QMLHFL} under increased model complexity, we also test it on the CIFAR-10 dataset using a convolutional neural network (CNN) comprising three convolutional blocks with 64, 128, and 256 filters, each containing two $3{\times}3$ ReLU-activated convolutional layers with batch normalization, followed by $2{\times}2$ max-pooling and dropout of 30\%, 40\%, and 40\%, respectively. The convolutional stack is followed by a dense layer of 512 ReLU-activated neurons with 50\% dropout and a 10-neuron softmax output layer. CIFAR-10 is evaluated only under the 6-layer setup. 
	
	We evaluate two configurations: (i) three sets of quantization levels for the lower layers $(s_1, s_2)$, while keeping $s_3 = 8$, $s_4 = 10$, $s_5 = 12$, and $s_6 = 14$ fixed; and (ii) three sets of quantization levels for the higher layers $(s_5, s_6)$, while fixing $s_1 = 4$, $s_2 = 6$, $s_3 = 8$, and $s_4 = 10$. For each configuration, the optimal intra-layer iteration counts are obtained using the proposed optimization framework under the timing parameters in Table~II and a deadline of $T_\text{d} = 3000T\ \text{sec}$. The resulting test accuracies after 100 global iterations are summarized in Tables~IV and~V for the three data heterogeneity cases.
	
	As observed, all configurations achieve satisfactory learning accuracy. However, increasing the quantization resolution in the lower layers yields a markedly larger performance gain, indicating that {\fontfamily{lmtt}\selectfont QMLHFL} is more sensitive to quantization precision in these layers. In contrast, while improving quantization resolution in the higher layers also enhances performance, its effect is significantly less pronounced, which is consistent with the insight discussed in \textit{Remark~9}.

	\vspace{-5pt}
	\section{Conclusions}
	In this work, we proposed a multi-layer hierarchical FL framework ({\fontfamily{lmtt}\selectfont QMLHFL}), which generalizes the traditional two-layer hierarchical FL framework to support architectures of arbitrary depth and configuration. It extends tractably through a nested aggregation structure, incorporating layer-specific iterations and applying layer-specific quantization due to communication constraints. We developed a comprehensive convergence analysis that provides theoretical guarantees, capturing the complex interactions across multiple aggregation layers. The analysis presents general convergence condition and rate expressions that explicitly highlight the impact of quantization, hierarchical architecture, nested aggregation structure, and intra-layer iteration counts---the main building blocks of {\fontfamily{lmtt}\selectfont QMLHFL}---along with several special cases. Grounded in this analysis, we formulated and solved a nonlinear optimization problem for tuning the iteration counts, enabling improved convergence under deadline constraints. Experimental results on multi-layer architectures demonstrate the effectiveness of {\fontfamily{lmtt}\selectfont QMLHFL}, both in terms of learning accuracy and runtime performance---particularly for the optimized variant.
	
	Building on the results of this work, future research can explore aggregation strategies beyond gradient descent, dynamic hierarchy formation, device selection, heterogeneity-aware analysis, asynchronous update mechanisms, and resource allocation.

	\newpage
	\appendix
	\section{Proof of Theorem 1}
	The global update of learning model at round $t+1$ is as
	\begin{align}
		&\mathbf{w}_{t+1} = \mathbf{w}_{t} -\mu \sum_{i_{N-1}}^{} \frac{|{\cal C}_{N-1}^{i_{N-1}}|}{N_\text{tot}}Q_N\Biggl(\sum_{t_N = 0}^{\tau_{N}-1}\sum_{i_{N-2} \in {\cal C}_{N-1}^{i_{N-1}}}^{} \frac{|{\cal C}_{N-2}^{i_{N-2}}|}{|{\cal C}_{N-1}^{i_{N-1}}|}\nonumber\\&Q_{N-1}\Biggl(\cdots \sum_{t_3=0}^{\tau_3-1}\sum_{i_1\in {\cal C}_2^{i_2}}^{} \frac{|{\cal C}_1^{i_1}|}{|{\cal C}_2^{i_2}|}Q_2\Biggl(\sum_{t_2=0}^{\tau_2-1}\sum_{i \in {\cal C}_1^{i_1}}^{}\frac{1}{|{\cal C}_1^{i_1}|}\nonumber\\&Q_1\Biggl(\sum_{t_1=0}^{\tau_1-1}\nabla F_{i}(\mathbf{w}_{t_1,\ldots,t_{N}}^{i,t},\boldsymbol\xi_{t_1,\ldots,t_{N}}^{i,t})\Biggr)\Biggr)\Biggr)\Biggr).
	\end{align}
	Subsequently, based on the $L$-Lipschitz inequality stated in Assumption 2 and applying expectation to its both sides, we obtain
	\begin{align}
		\label{lipexpp}
		&\mathbb{E}\left\{F({\mathbf{w}}_{t+1})-F({\mathbf{w}}_{t})\right\} \leq \mathbb{E}\left\{\nabla F( \mathbf{w}_{t})^\top \left({\mathbf{w}}_{t+1}- {\mathbf{w}}_{t}\right)\right\}\nonumber\\&+\frac{L}{2} \mathbb{E}\left\{\|{\mathbf{w}}_{t+1} - {\mathbf{w}}_{t}\|^2\right\},
	\end{align}
	which plays the key role in the proof, as will be shown in~\eqref{lipexp}-\eqref{finnal_sttep}. Subsequently, we derive the terms in the right side in~\eqref{lipexpp}, which constitute a substantial portion of the proof. 
	
	\textbf{First Term of~\eqref{lipexpp}:} It is obtained as
	\begin{align}
		\label{main1}
		&\mathbb{E}\left\{\nabla F( \mathbf{w}_{t})^\top \left({\mathbf{w}}_{t+1}- {\mathbf{w}}_{t}\right)\right\} = -\mu \sum_{i_{N-1}}^{} \frac{|{\cal C}_{N-1}^{i_{N-1}}|}{N_\text{tot}}\sum_{t_N = 0}^{\tau_{N}-1}\nonumber\\&\sum_{i_{N-2} \in {\cal C}_{N-1}^{i_{N-1}}}^{} \frac{|{\cal C}_{N-2}^{i_{N-2}}|}{|{\cal C}_{N-1}^{i_{N-1}}|}\cdots \sum_{t_3=0}^{\tau_3-1}\sum_{i_1\in {\cal C}_2^{i_2}}^{} \frac{|{\cal C}_1^{i_1}|}{|{\cal C}_2^{i_2}|}\sum_{t_2=0}^{\tau_2-1}\sum_{i \in {\cal C}_1^{i_1}}^{}\frac{1}{|{\cal C}_1^{i_1}|}\nonumber\\&\sum_{t_1=0}^{\tau_1-1}\mathbb{E}\left\{\nabla F( \mathbf{w}_{t})^\top\nabla F_{i}(\mathbf{w}_{t_1,\ldots,t_{N}}^{i,t},\boldsymbol\xi_{t_1,\ldots,t_{N}}^{i,t})\right\} = -\frac{\mu}{N_\text{tot}}\nonumber\\&\sum_{i_{N-1}}^{} \sum_{t_N = 0}^{\tau_{N}-1}\sum_{i_{N-2} \in {\cal C}_{N-1}^{i_{N-1}}}^{}\cdots \sum_{t_3=0}^{\tau_3-1}\sum_{i_1\in {\cal C}_2^{i_2}}^{}\sum_{t_2=0}^{\tau_2-1}\sum_{i \in {\cal C}_1^{i_1}}^{}\sum_{t_1=0}^{\tau_1-1}\nonumber\\&\mathbb{E}\left\{\nabla F( \mathbf{w}_{t})^\top\nabla F_{i}(\mathbf{w}_{t_1,\ldots,t_{N}}^{i,t},\boldsymbol\xi_{t_1,\ldots,t_{N}}^{i,t})\right\} = -\frac{\mu}{N_\text{tot}}\sum_{i}^{} \sum_{t_N = 0}^{\tau_{N}-1}\nonumber\\&\cdots \sum_{t_3=0}^{\tau_3-1}\sum_{t_2=0}^{\tau_2-1}\sum_{t_1=0}^{\tau_1-1}\mathbb{E}\left\{\nabla F( \mathbf{w}_{t})^\top\nabla F_{i}(\mathbf{w}_{t_1,\ldots,t_{N}}^{i,t},\boldsymbol\xi_{t_1,\ldots,t_{N}}^{i,t})\right\} \nonumber\\&=-\frac{\mu}{N_\text{tot}}\sum_{i}^{} \sum_{t_N = 0}^{\tau_{N}-1}\cdots \sum_{t_3=0}^{\tau_3-1}\sum_{t_2=0}^{\tau_2-1}\sum_{t_1=0}^{\tau_1-1}\nonumber\\&\mathbb{E}\left\{\nabla F( \mathbf{w}_{t})^\top\nabla F(\mathbf{w}_{t_1,\ldots,t_{N}}^{i,t})\right\},
	\end{align}
	where, by employing the equation $\|\mathbf{t}_1-\mathbf{t}_2\|^2 = \|\mathbf{t}_1\|^2+\|\mathbf{t}_2\|^2- 2\mathbf{t}_1^\top \mathbf{t}_2$ for any vectors $\mathbf{t}_1$ and $\mathbf{t}_2$, the inner-sum term is expressed as
	\begin{align}
		&\label{diffexp}
		\mathbb{E}\left\{\nabla F( {\mathbf{w}}_{t})^\top \nabla F(\mathbf{w}_{t_1,\ldots,t_{N}}^{i,t})\right\} = \frac{1}{2} \mathbb{E}\left\{\|\nabla F( {\mathbf{w}}_{t})\|^2\right\} \nonumber\\&+ \frac{1}{2} \mathbb{E}\left\{\|\nabla F(\mathbf{w}_{t_1,\ldots,t_{N}}^{i,t})\|^2\right\} \nonumber\\&- \frac{1}{2} \mathbb{E}\left\{\|\nabla F( {\mathbf{w}}_{t})- \nabla F(\mathbf{w}_{t_1,\ldots,t_{N}}^{i,t})\|^2\right\}.
	\end{align}
	From Assumption 2, \underline{the last term in the right side of~\eqref{diffexp}} is bounded as
	\begin{align}
		\label{difference}
		&\mathbb{E}\left\{\|\nabla F( {\mathbf{w}}_{t})- \nabla F(\mathbf{w}_{t_1,\ldots,t_{N}}^{i,t})\|^2\right\} \leq \nonumber\\
		&L^2 \mathbb{E}\left\{\|\mathbf{w}_{t_1,\ldots,t_{N}}^{i,t} - \mathbf{w}_t\|^2\right\}
		= L^2\mu^2 \mathbb{E}\Biggl\{\Biggl\Vert\sum_{r_1=0}^{t_1-1}\nonumber\\&\nabla F_i(\mathbf{w}_{r_1,\ldots,t_{N}}^{i,t}, \boldsymbol\xi_{r_1,\ldots,t_N}^{i,t}) + \sum_{r_2=0}^{t_2-1}\sum_{j \in {\cal C}_{1,i}}^{}\frac{1}{|{\cal C}_{1,i}|}Q_1\Biggl(\sum_{r_1=0}^{\tau_1-1}\nonumber\\&\nabla F_{j}(\mathbf{w}_{r_1, r_2,\ldots,t_{N}}^{j,t},\boldsymbol\xi_{r_1,r_2,\ldots,t_{N}}^{j,t})\Biggr)+\sum_{r_3=0}^{t_3-1}\sum_{j_1\in {\cal C}_{2,i}}^{} \frac{|{\cal C}_{1}^{j_1}|}{|{\cal C}_{2,i}|}\nonumber\\&Q_2\Biggl(\sum_{r_2=0}^{\tau_2-1}\sum_{j \in {\cal C}_{1}^{j_1}}^{}\frac{1}{|{\cal C}_{1}^{j_1}|}Q_1\Biggl(\sum_{r_1=0}^{\tau_1-1}\nabla F_{j}(\mathbf{w}_{r_1,\ldots,t_{N}}^{j,t},\boldsymbol\xi_{r_1,\ldots,t_{N}}^{j,t})\Biggr)\Biggr)\nonumber\\&+\cdots+\sum_{r_N = 0}^{t_{N}-1}\sum_{j_{N-2} \in {\cal C}_{N-1,i}}^{} \frac{|{\cal C}_{N-2}^{j_{N-2}}|}{|{\cal C}_{N-1,i}|}Q_{N-1}\Biggl(\cdots \sum_{r_3=0}^{\tau_3-1}\sum_{j_1\in {\cal C}_2^{j_2}}^{} \frac{|{\cal C}_1^{j_1}|}{|{\cal C}_2^{j_2}|}\nonumber\\&Q_2\Biggl(\sum_{r_2=0}^{\tau_2-1}\sum_{j \in {\cal C}_1^{j_1}}^{}\frac{1}{|{\cal C}_1^{j_1}|}Q_1\Biggl(\sum_{r_1=0}^{\tau_1-1}\nabla F_{j}(\mathbf{w}_{r_1,\ldots,r_{N}}^{j,t},\boldsymbol\xi_{r_1,\ldots,r_{N}}^{j,t})\Biggr)\Biggr)\Biggr)\nonumber\\&\Biggr\Vert^2\Biggr\}\stackrel{(a)}{=} L^2\mu^2 \mathbb{E}\Biggl\{\Biggl\Vert\sum_{r_1=0}^{t_1-1}\nabla F(\mathbf{w}_{r_1,\ldots,t_{N}}^{i,t}) + \sum_{j \in {\cal C}_{1,i}}^{}\frac{1}{|{\cal C}_{1,i}|}\sum_{r_2=0}^{t_2-1}\nonumber\\&\sum_{r_1=0}^{\tau_1-1}\nabla F(\mathbf{w}_{r_1, r_2,\ldots,t_{N}}^{j,t})+\sum_{j_1\in {\cal C}_{2,i}}^{}\sum_{j \in {\cal C}_{1}^{j_1}}^{} \frac{1}{|{\cal C}_{2,i}|}\sum_{r_3=0}^{t_3-1}\sum_{r_2=0}^{\tau_2-1}\sum_{r_1=0}^{\tau_1-1}\nonumber\\&\nabla F(\mathbf{w}_{r_1,\ldots,t_{N}}^{j,t})+\cdots+\sum_{j_{N-2} \in {\cal C}_{N-1,i}}^{} \cdots \sum_{j \in {\cal C}_1^{j_1}}^{}\frac{1}{|{\cal C}_{N-1,i}|}\sum_{r_N = 0}^{t_{N}-1}\nonumber\\&\cdots  \sum_{r_2=0}^{\tau_2-1}\sum_{r_1=0}^{\tau_1-1}\nabla F(\mathbf{w}_{r_1,\ldots,r_{N}}^{j,t})\Biggr\Vert^2\Biggr\}+L^2\mu^2 \mathbb{E}\Biggl\{\Biggl\Vert\sum_{r_1=0}^{t_1-1}\nonumber\\&\nabla F_i(\mathbf{w}_{r_1,\ldots,t_{N}}^{i,t}, \boldsymbol\xi_{r_1,\ldots,t_N}^{i,t}) - \nabla F(\mathbf{w}_{r_1,\ldots,t_{N}}^{i,t}) \Biggr\Vert^2\Biggr\}+ L^2\mu^2\nonumber\\& \mathbb{E}\Biggl\{\Biggl\Vert \sum_{j \in {\cal C}_{1,i}}^{}\frac{1}{|{\cal C}_{1,i}|}\sum_{r_2=0}^{t_2-1}\Biggl[Q_1\Biggl(\sum_{r_1=0}^{\tau_1-1}\nabla F_{j}(\mathbf{w}_{r_1, r_2,\ldots,t_{N}}^{j,t},\boldsymbol\xi_{r_1,r_2,\ldots,t_{N}}^{j,t})\nonumber\\&\Biggr) - \sum_{r_1=0}^{\tau_1-1}\nabla F(\mathbf{w}_{r_1, r_2,\ldots,t_{N}}^{j,t})\Biggr]\Biggr\Vert^2 \Biggr\}+L^2\mu^2 \mathbb{E}\Biggl\{\Biggl\Vert \sum_{j_1\in {\cal C}_{2,i}}^{} \frac{|{\cal C}_{1}^{j_1}|}{|{\cal C}_{2,i}|}\sum_{r_3=0}^{t_3-1}\nonumber\\&\Biggl[Q_2\Biggl(\sum_{r_2=0}^{\tau_2-1}\sum_{j \in {\cal C}_{1}^{j_1}}^{}\frac{1}{|{\cal C}_{1}^{j_1}|}Q_1\Biggl(\sum_{r_1=0}^{\tau_1-1}\nabla F_{j}(\mathbf{w}_{r_1,\ldots,t_{N}}^{j,t},\boldsymbol\xi_{r_1,\ldots,t_{N}}^{j,t})\Biggr)\Biggr) \nonumber\\&- \sum_{r_2=0}^{\tau_2-1}\sum_{j \in {\cal C}_{1}^{j_1}}^{}\frac{1}{|{\cal C}_{1}^{j_1}|}\sum_{r_1=0}^{\tau_1-1}\nabla F(\mathbf{w}_{r_1,\ldots,t_{N}}^{j,t})  \Biggr]\Biggr\Vert^2\Biggr\}+\cdots +L^2\mu^2 \nonumber\\&\mathbb{E}\Biggl\{\Biggl\Vert \sum_{j_{N-2} \in {\cal C}_{N-1,i}}^{} \frac{|{\cal C}_{N-2}^{j_{N-2}}|}{|{\cal C}_{N-1,i}|}\sum_{r_N = 0}^{t_{N}-1}Q_{N-1}\Biggl(\cdots \sum_{r_3=0}^{\tau_3-1}\sum_{j_1\in {\cal C}_2^{j_2}}^{} \frac{|{\cal C}_1^{j_1}|}{|{\cal C}_2^{j_2}|}\nonumber\\&Q_2\Biggl(\sum_{r_2=0}^{\tau_2-1}\sum_{j \in {\cal C}_1^{j_1}}^{}\frac{1}{|{\cal C}_1^{j_1}|}Q_1\Biggl(\sum_{r_1=0}^{\tau_1-1}\nabla F_{j}(\mathbf{w}_{r_1,\ldots,r_{N}}^{j,t},\boldsymbol\xi_{r_1,\ldots,r_{N}}^{j,t})\Biggr)\Biggr)\Biggr) \nonumber
	\end{align}
	\begin{align}
		&-\cdots \sum_{r_3=0}^{\tau_3-1}\sum_{j_1\in {\cal C}_2^{j_2}}^{} \frac{|{\cal C}_1^{j_1}|}{|{\cal C}_2^{j_2}|}\sum_{r_2=0}^{\tau_2-1}\sum_{j \in {\cal C}_1^{j_1}}^{}\frac{1}{|{\cal C}_1^{j_1}|}\sum_{r_1=0}^{\tau_1-1}\nabla F(\mathbf{w}_{r_1,\ldots,r_{N}}^{j,t})\Biggr\Vert^2\nonumber\\
		&\Biggr\},
	\end{align}
	where $(a)$ comes from the identity $\mathbb{E}\left\{\|\mathbf{t}\|^2\right\} = \|\mathbb{E}\left\{\mathbf{t}\right\}\|^2 + \mathbb{E}\left\{\|\mathbf{t}-\mathbb{E}\left\{\mathbf{t}\right\}\|^2\right\}$ for any vector $\mathbf{t}$. Also, ${\cal C}_{n,i}, \forall n= 1, \ldots, N-1$, represents the set of edge servers at layer $n-1$ that are connected to the same edge server at layer $n$, with one of them hierarchically connected to the $i$-th device. The notation $|{\cal C}_{n,i}|$ indicates the number of devices connected to ${\cal C}_{n,i}$. 
	
	\underline{The first term in the right side of~\eqref{difference}} is bounded as
	\begin{align}
		\label{purenorm}
		&\mathbb{E}\Biggl\{\Biggl\Vert\sum_{r_1=0}^{t_1-1}\nabla F(\mathbf{w}_{r_1,\ldots,t_{N}}^{i,t}) + \sum_{j \in {\cal C}_{1,i}}^{}\frac{1}{|{\cal C}_{1,i}|}\sum_{r_2=0}^{t_2-1}\sum_{r_1=0}^{\tau_1-1}\nonumber\\&\nabla F(\mathbf{w}_{r_1, r_2,\ldots,t_{N}}^{j,t})+\sum_{j_1\in {\cal C}_{2,i}}^{}\sum_{j \in {\cal C}_{1}^{j_1}}^{} \frac{1}{|{\cal C}_{2,i}|}\sum_{r_3=0}^{t_3-1}\sum_{r_2=0}^{\tau_2-1}\sum_{r_1=0}^{\tau_1-1}\nonumber\\&\nabla F(\mathbf{w}_{r_1,\ldots,t_{N}}^{j,t})+\cdots+\sum_{j_{N-2} \in {\cal C}_{N-1,i}}^{} \cdots \sum_{j \in {\cal C}_1^{j_1}}^{}\frac{1}{|{\cal C}_{N-1,i}|}\nonumber\\&\sum_{r_N = 0}^{t_{N}-1}\cdots  \sum_{r_2=0}^{\tau_2-1}\sum_{r_1=0}^{\tau_1-1}\nabla F(\mathbf{w}_{r_1,\ldots,r_{N}}^{j,t})\Biggr\Vert^2\Biggr\} \stackrel{(b)}{\leq} t_1\sum_{r_1=0}^{t_1-1}\nonumber\\&\mathbb{E}\Biggl\{\Biggl\Vert\nabla F(\mathbf{w}_{r_1,\ldots,t_{N}}^{i,t})\Biggr\Vert^2\Biggr\}+ t_2 \tau_1 \frac{1}{|{\cal C}_{1,i}|} \sum_{j \in {\cal C}_{1,i}}^{}\sum_{r_2=0}^{t_2-1}\sum_{r_1=0}^{\tau_1-1}\nonumber\\&\mathbb{E}\Biggl\{\Biggl\Vert\nabla F(\mathbf{w}_{r_1, r_2,\ldots,t_{N}}^{j,t})\Biggr\Vert^2\Biggr\} + t_3 \tau_2 \tau_1 \frac{1}{|{\cal C}_{2,i}|}\sum_{j_1\in {\cal C}_{2,i}}^{}\sum_{j \in {\cal C}_{1}^{j_1}}^{} \sum_{r_3=0}^{t_3-1}\nonumber\\&\sum_{r_2=0}^{\tau_2-1}\sum_{r_1=0}^{\tau_1-1}\mathbb{E}\Biggl\{\Biggl\Vert\nabla F(\mathbf{w}_{r_1,\ldots,t_{N}}^{j,t})\Biggr\Vert^2\Biggr\}+\cdots+t_N \tau_{N-1} \cdots \tau_1 \nonumber\\&\frac{1}{|{\cal C}_{N-1,i}|} \sum_{j_{N-2} \in {\cal C}_{N-1,i}}^{} \cdots \sum_{j \in {\cal C}_1^{j_1}}^{}\sum_{r_N = 0}^{t_{N}-1}\cdots  \sum_{r_2=0}^{\tau_2-1}\sum_{r_1=0}^{\tau_1-1}\nonumber\\&\mathbb{E}\Biggl\{\Biggl\Vert\nabla F(\mathbf{w}_{r_1,\ldots,r_{N}}^{j,t})\Biggr\Vert^2\Biggr\},
	\end{align}
	where $(b)$ follows from the arithmetic-geometric mean inequality, given by \(
	\left(\sum_{j=1}^{J}t_j\right)^2 \leq J \sum_{j=1}^{J}t_j^2.\) For \underline{the second term of~\eqref{difference}}, we have
	\begin{align}
		\label{2th}
		&\mathbb{E}\Biggl\{\Biggl\Vert\sum_{r_1=0}^{t_1-1}\nabla F_i(\mathbf{w}_{r_1,\ldots,t_{N}}^{i,t}, \boldsymbol\xi_{r_1,\ldots,t_N}^{i,t}) - \nabla F(\mathbf{w}_{r_1,\ldots,t_{N}}^{i,t}) \Biggr\Vert^2\Biggr\} \nonumber\\&=  \sum_{r_1=0}^{t_1-1}\mathbb{E}\Biggl\{\Biggl\Vert\nabla F_i(\mathbf{w}_{r_1,\ldots,t_{N}}^{i,t}, \boldsymbol\xi_{r_1,\ldots,t_N}^{i,t}) - \nabla F(\mathbf{w}_{r_1,\ldots,t_{N}}^{i,t}) \Biggr\Vert^2\Biggr\}\nonumber\\&\leq t_1 \sigma^2,
	\end{align}
	which is from Assumption 3. For \underline{the third term of~\eqref{difference}}, we have
	\begin{align}
		\label{3th}
		&\mathbb{E}\Biggl\{\Biggl\Vert \sum_{j \in {\cal C}_{1,i}}^{}\frac{1}{|{\cal C}_{1,i}|}\sum_{r_2=0}^{t_2-1}\Biggl[Q_1\Biggl(\sum_{r_1=0}^{\tau_1-1}\nabla F_{j}(\mathbf{w}_{r_1, r_2,\ldots,t_{N}}^{j,t},\boldsymbol\xi_{r_1,r_2,\ldots,t_{N}}^{j,t}) \nonumber\\&\Biggr)- \sum_{r_1=0}^{\tau_1-1}\nabla F(\mathbf{w}_{r_1, r_2,\ldots,t_{N}}^{j,t})\Biggr]\Biggr\Vert^2 \Biggr\}  = \frac{1}{|{\cal C}_{1,i}|^2} \sum_{j \in {\cal C}_{1,i}}^{}\sum_{r_2=0}^{t_2-1}\nonumber\\&\mathbb{E}\Biggl\{\Biggl\Vert Q_1\Biggl(\sum_{r_1=0}^{\tau_1-1}\nabla F_{j}(\mathbf{w}_{r_1, r_2,\ldots,t_{N}}^{j,t},\boldsymbol\xi_{r_1,r_2,\ldots,t_{N}}^{j,t})\Biggr) - \sum_{r_1=0}^{\tau_1-1}\nonumber\\
		&\nabla F(\mathbf{w}_{r_1, r_2,\ldots,t_{N}}^{j,t})\Biggr\Vert^2 \Biggr\} \stackrel{(c)}{=}\frac{1}{|{\cal C}_{1,i}|^2} \sum_{j \in {\cal C}_{1,i}}^{}\sum_{r_2=0}^{t_2-1}\Biggl[\tau_1\sigma^2 + q_1\nonumber\\& \mathbb{E}\Biggl\{\Biggl\Vert\sum_{r_1=0}^{\tau_1-1}\nabla F_{j}(\mathbf{w}_{r_1, r_2,\ldots,t_{N}}^{j,t},\boldsymbol\xi_{r_1,r_2,\ldots,t_{N}}^{j,t})\Biggr\Vert^2\Biggr\} \Biggr]  =\frac{1}{|{\cal C}_{1,i}|^2} \sum_{j \in {\cal C}_{1,i}}^{}\nonumber\\&\sum_{r_2=0}^{t_2-1}\Biggl[\tau_1\sigma^2 + q_1 \mathbb{E}\Biggl\{\Biggl\Vert\sum_{r_1=0}^{\tau_1-1}\nabla F(\mathbf{w}_{r_1, r_2,\ldots,t_{N}}^{j,t})\Biggr\Vert^2\Biggr\}+q_1\tau_1\sigma^2 \Biggr] \nonumber\\&\leq (1+q_1)  t_2 \tau_1 \sigma^2 \frac{1}{|{\cal C}_{1,i}|} + \frac{1}{|{\cal C}_{1,i}|^2} q_1\tau_1 \sum_{j \in {\cal C}_{1,i}}^{}\sum_{r_2=0}^{t_2-1}\sum_{r_1=0}^{\tau_1-1}\nonumber\\&\mathbb{E}\Biggl\{\Biggl\Vert\nabla F(\mathbf{w}_{r_1, r_2,\ldots,t_{N}}^{j,t})\Biggr\Vert^2\Biggr\},
	\end{align}
	where $(c)$ is from Assumption 1. Next, for \underline{the forth term of the right side of~\eqref{difference}}, we have
	\begin{align}
		\label{4th}
		&\mathbb{E}\Biggl\{\Biggl\Vert \sum_{j_1\in {\cal C}_{2,i}}^{} \frac{|{\cal C}_{1}^{j_1}|}{|{\cal C}_{2,i}|}\sum_{r_3=0}^{t_3-1}\Biggl[Q_2\Biggl(\sum_{r_2=0}^{\tau_2-1}\sum_{j \in {\cal C}_{1}^{j_1}}^{}\frac{1}{|{\cal C}_{1}^{j_1}|}Q_1\Biggl(\sum_{r_1=0}^{\tau_1-1}\nonumber\\&\nabla F_{j}(\mathbf{w}_{r_1,\ldots,t_{N}}^{j,t},\boldsymbol\xi_{r_1,\ldots,t_{N}}^{j,t})\Biggr)\Biggr) - \sum_{r_2=0}^{\tau_2-1}\sum_{j \in {\cal C}_{1}^{j_1}}^{}\frac{1}{|{\cal C}_{1}^{j_1}|}\sum_{r_1=0}^{\tau_1-1}\nonumber\\&\nabla F(\mathbf{w}_{r_1,\ldots,t_{N}}^{j,t})  \Biggr]\Biggr\Vert^2\Biggr\}  =  \frac{1}{|{\cal C}_{2,i}|^2}\sum_{j_1\in {\cal C}_{2,i}}^{} {|{\cal C}_{1}^{j_1}|^2}\sum_{r_3=0}^{t_3-1}\mathbb{E}\Biggl\{\Biggl\Vert\nonumber\\& Q_2\Biggl(\sum_{r_2=0}^{\tau_2-1}\sum_{j \in {\cal C}_{1}^{j_1}}^{}\frac{1}{|{\cal C}_{1}^{j_1}|}Q_1\Biggl(\sum_{r_1=0}^{\tau_1-1}\nabla F_{j}(\mathbf{w}_{r_1,\ldots,t_{N}}^{j,t},\boldsymbol\xi_{r_1,\ldots,t_{N}}^{j,t})\Biggr)\Biggr) \nonumber\\&- \sum_{r_2=0}^{\tau_2-1}\sum_{j \in {\cal C}_{1}^{j_1}}^{}\frac{1}{|{\cal C}_{1}^{j_1}|}\sum_{r_1=0}^{\tau_1-1}\nabla F(\mathbf{w}_{r_1,\ldots,t_{N}}^{j,t})  \Biggl\Vert^2\Biggr\} = \frac{1}{|{\cal C}_{2,i}|^2}\sum_{j_1\in {\cal C}_{2,i}}^{} \nonumber\\&{|{\cal C}_{1}^{j_1}|^2}\sum_{r_3=0}^{t_3-1}\Biggl[\mathbb{E}\Biggl\{\Biggl\Vert Q_2\Biggl(\sum_{r_2=0}^{\tau_2-1}\sum_{j \in {\cal C}_{1}^{j_1}}^{}\frac{1}{|{\cal C}_{1}^{j_1}|}Q_1\Biggl(\sum_{r_1=0}^{\tau_1-1}\nonumber\\&\nabla F_{j}(\mathbf{w}_{r_1,\ldots,t_{N}}^{j,t},\boldsymbol\xi_{r_1,\ldots,t_{N}}^{j,t})\Biggr)\Biggr) - \sum_{r_2=0}^{\tau_2-1}\sum_{j \in {\cal C}_{1}^{j_1}}^{}\frac{1}{|{\cal C}_{1}^{j_1}|}Q_1\Biggl(\sum_{r_1=0}^{\tau_1-1}\nonumber\\&\nabla F_{j}(\mathbf{w}_{r_1,\ldots,t_{N}}^{j,t},\boldsymbol\xi_{r_1,\ldots,t_{N}}^{j,t})\Biggr)\Biggr\Vert^2\Biggr\}+\mathbb{E}\Biggl\{\Biggl\Vert\sum_{r_2=0}^{\tau_2-1}\sum_{j \in {\cal C}_{1}^{j_1}}^{}\frac{1}{|{\cal C}_{1}^{j_1}|}\nonumber
	\end{align}
	\begin{align}
		&Q_1\Biggl(\sum_{r_1=0}^{\tau_1-1}\nabla F_{j}(\mathbf{w}_{r_1,\ldots,t_{N}}^{j,t},\boldsymbol\xi_{r_1,\ldots,t_{N}}^{j,t})\Biggr) -\sum_{r_2=0}^{\tau_2-1}\sum_{j \in {\cal C}_{1}^{j_1}}^{}\frac{1}{|{\cal C}_{1}^{j_1}|}\nonumber\\
		&\sum_{r_1=0}^{\tau_1-1}\nabla F(\mathbf{w}_{r_1,\ldots,t_{N}}^{j,t})\Biggr\Vert^2\Biggr\}\Biggr] =\frac{1}{|{\cal C}_{2,i}|^2}\sum_{j_1\in {\cal C}_{2,i}}^{} {|{\cal C}_{1}^{j_1}|^2}\sum_{r_3=0}^{t_3-1}\nonumber\\&\Biggl[q_2\mathbb{E}\Biggl\{\Biggl\Vert \sum_{r_2=0}^{\tau_2-1}\sum_{j \in {\cal C}_{1}^{j_1}}^{}\frac{1}{|{\cal C}_{1}^{j_1}|}Q_1\Biggl(\sum_{r_1=0}^{\tau_1-1}\nabla F_{j}(\mathbf{w}_{r_1,\ldots,t_{N}}^{j,t},\boldsymbol\xi_{r_1,\ldots,t_{N}}^{j,t})\nonumber\\&\Biggr)\Biggr\Vert^2\Biggr\}+\sum_{r_2=0}^{\tau_2-1}\sum_{j \in {\cal C}_{1}^{j_1}}^{}\frac{1}{|{\cal C}_{1}^{j_1}|^2}\mathbb{E}\Biggl\{\Biggl\Vert \nonumber\\
		&Q_1\Biggl(\sum_{r_1=0}^{\tau_1-1}\nabla F_{j}(\mathbf{w}_{r_1,\ldots,t_{N}}^{j,t},\boldsymbol\xi_{r_1,\ldots,t_{N}}^{j,t})\Biggr) -\sum_{r_1=0}^{\tau_1-1}\nabla F(\mathbf{w}_{r_1,\ldots,t_{N}}^{j,t})\nonumber\\
		&\Biggr\Vert^2\Biggr\}\Biggr]  = \frac{1}{|{\cal C}_{2,i}|^2}\sum_{j_1\in {\cal C}_{2,i}}^{} {|{\cal C}_{1}^{j_1}|^2}\sum_{r_3=0}^{t_3-1}\Biggl[q_2\mathbb{E}\Biggl\{\Biggl\Vert \sum_{r_2=0}^{\tau_2-1}\sum_{j \in {\cal C}_{1}^{j_1}}^{}\frac{1}{|{\cal C}_{1}^{j_1}|}\nonumber\\
		&\sum_{r_1=0}^{\tau_1-1}\nabla F(\mathbf{w}_{r_1,\ldots,t_{N}}^{j,t})\Biggr\Vert^2\Biggr\}+(1+q_2)\sum_{r_2=0}^{\tau_2-1}\sum_{j \in {\cal C}_{1}^{j_1}}^{}\frac{1}{|{\cal C}_{1}^{j_1}|^2}\mathbb{E}\Biggl\{\Biggl\Vert\nonumber\\
		& Q_1\Biggl(\sum_{r_1=0}^{\tau_1-1}\nabla F_{j}(\mathbf{w}_{r_1,\ldots,t_{N}}^{j,t},\boldsymbol\xi_{r_1,\ldots,t_{N}}^{j,t})\Biggr) -\sum_{r_1=0}^{\tau_1-1}\nonumber\\&\nabla F(\mathbf{w}_{r_1,\ldots,t_{N}}^{j,t})\Biggr\Vert^2\Biggr\}\Biggr]\leq \frac{1}{|{\cal C}_{2,i}|^2}\sum_{j_1\in {\cal C}_{2,i}}^{} {|{\cal C}_{1}^{j_1}|^2}\sum_{r_3=0}^{t_3-1}\Biggl[q_2\tau_1 \tau_2\nonumber\\&\frac{1}{|{\cal C}_{1}^{j_1}|}\sum_{j \in {\cal C}_{1}^{j_1}}^{}  \sum_{r_2=0}^{\tau_2-1}\sum_{r_1=0}^{\tau_1-1}\mathbb{E}\Biggl\{\Biggl\Vert\nabla F(\mathbf{w}_{r_1,\ldots,t_{N}}^{j,t})\Biggr\Vert^2\Biggr\}+(1+q_2)\nonumber\\&\frac{1}{|{\cal C}_{1}^{j_1}|^2}\sum_{r_2=0}^{\tau_2-1}\sum_{j \in {\cal C}_{1}^{j_1}}^{} \Biggl[(1+q_1)\tau_1\sigma^2 + q_1 \tau_1 \sum_{r_1=0}^{\tau_1-1}\mathbb{E}\Biggl\{\Biggl\Vert\nonumber\\&\nabla F(\mathbf{w}_{r_1,\ldots,t_{N}}^{j,t})\Biggr\Vert^2\Biggr\} \Biggr]\Biggr]  = (1+q_2)(1+q_1)t_3\tau_2 \tau_1\sigma^2\frac{1}{|{\cal C}_{2,i}|}  \nonumber\\&+ \frac{1}{|{\cal C}_{2,i}|^2}\sum_{j_1\in {\cal C}_{2,i}}^{} {|{\cal C}_{1}^{j_1}|}\left(q_2 \tau_1 \tau_2+\frac{1}{|{\cal C}_{1}^{j_1}|} q_1 (1+q_2)\tau_1 \right)\sum_{j \in {\cal C}_{1}^{j_1}}^{}\nonumber\\&\sum_{r_3=0}^{t_3-1}  \sum_{r_2=0}^{\tau_2-1}\sum_{r_1=0}^{\tau_1-1}\mathbb{E}\Biggl\{\Biggl\Vert\nabla F(\mathbf{w}_{r_1,\ldots,t_{N}}^{j,t})\Biggr\Vert^2\Biggr\},
	\end{align}
	and for \underline{the fifth term of the right side of~\eqref{difference}}, we have
	\begin{align}
		\label{5th}
		&\mathbb{E}\Biggl\{\Biggl\Vert \sum_{j_{2} \in {\cal C}_{3,i}}^{} \frac{|{\cal C}_{2}^{j_{2}}|}{|{\cal C}_{3,i}|}\sum_{r_4 = 0}^{t_{4}-1}Q_{3}\Biggl( \sum_{r_3=0}^{\tau_3-1}\sum_{j_1\in {\cal C}_2^{j_2}}^{} \frac{|{\cal C}_1^{j_1}|}{|{\cal C}_2^{j_2}|}Q_2\Biggl(\sum_{r_2=0}^{\tau_2-1}\sum_{j \in {\cal C}_1^{j_1}}^{}\nonumber\\&\frac{1}{|{\cal C}_1^{j_1}|}Q_1\Biggl(\sum_{r_1=0}^{\tau_1-1}\nabla F_{j}(\mathbf{w}_{r_1,\ldots,r_{N}}^{j,t},\boldsymbol\xi_{r_1,\ldots,r_{N}}^{j,t})\Biggr)\Biggr)\Biggr) - \sum_{r_3=0}^{\tau_3-1}\sum_{j_1\in {\cal C}_2^{j_2}}^{} \nonumber\\&\frac{|{\cal C}_1^{j_1}|}{|{\cal C}_2^{j_2}|}\sum_{r_2=0}^{\tau_2-1}\sum_{j \in {\cal C}_1^{j_1}}^{}\frac{1}{|{\cal C}_1^{j_1}|}\sum_{r_1=0}^{\tau_1-1}\nabla F(\mathbf{w}_{r_1,\ldots,r_{N}}^{j,t})\Biggr\Vert^2\Biggr\} = \sum_{j_{2} \in {\cal C}_{3,i}}^{} \frac{|{\cal C}_{2}^{j_{2}}|^2}{|{\cal C}_{3,i}|^2}\nonumber
	\end{align}
	\begin{align}
		&\sum_{r_4 = 0}^{t_{4}-1}\mathbb{E}\Biggl\{\Biggl\Vert Q_{3}\Biggl( \sum_{r_3=0}^{\tau_3-1}\sum_{j_1\in {\cal C}_2^{j_2}}^{} \frac{|{\cal C}_1^{j_1}|}{|{\cal C}_2^{j_2}|}Q_2\Biggl(\sum_{r_2=0}^{\tau_2-1}\sum_{j \in {\cal C}_1^{j_1}}^{}\frac{1}{|{\cal C}_1^{j_1}|}Q_1\Biggl(\nonumber\\
		&\sum_{r_1=0}^{\tau_1-1}\nabla F_{j}(\mathbf{w}_{r_1,\ldots,r_{N}}^{j,t},\boldsymbol\xi_{r_1,\ldots,r_{N}}^{j,t})\Biggr)\Biggr)\Biggr) - \sum_{r_3=0}^{\tau_3-1}\sum_{j_1\in {\cal C}_2^{j_2}}^{} \frac{|{\cal C}_1^{j_1}|}{|{\cal C}_2^{j_2}|}\sum_{r_2=0}^{\tau_2-1}\nonumber\\
		&\sum_{j \in {\cal C}_1^{j_1}}^{}\frac{1}{|{\cal C}_1^{j_1}|}\sum_{r_1=0}^{\tau_1-1}\nabla F(\mathbf{w}_{r_1,\ldots,r_{N}}^{j,t})\Biggr\Vert^2\Biggr\} = \sum_{j_{2} \in {\cal C}_{3,i}}^{} \frac{|{\cal C}_{2}^{j_{2}}|^2}{|{\cal C}_{3,i}|^2}\sum_{r_4 = 0}^{t_{4}-1}\Biggl[q_3\nonumber\\
		&\mathbb{E}\Biggl\{\Biggl\Vert \sum_{r_3=0}^{\tau_3-1}\sum_{j_1\in {\cal C}_2^{j_2}}^{} \frac{|{\cal C}_1^{j_1}|}{|{\cal C}_2^{j_2}|}Q_2\Biggl(\sum_{r_2=0}^{\tau_2-1}\sum_{j \in {\cal C}_1^{j_1}}^{}\frac{1}{|{\cal C}_1^{j_1}|}Q_1\Biggl(\sum_{r_1=0}^{\tau_1-1}\nonumber\\
		&\nabla F_{j}(\mathbf{w}_{r_1,\ldots,r_{N}}^{j,t},\boldsymbol\xi_{r_1,\ldots,r_{N}}^{j,t})\Biggr)\Biggr)\Biggr\Vert^2 \Biggr\}+ \sum_{r_3=0}^{\tau_3-1}\sum_{j_1\in {\cal C}_2^{j_2}}^{} \frac{|{\cal C}_1^{j_1}|^2}{|{\cal C}_2^{j_2}|^2}\mathbb{E}\Biggl\{\nonumber\\
		&\Biggl\Vert Q_2\Biggl(\sum_{r_2=0}^{\tau_2-1}\sum_{j \in {\cal C}_1^{j_1}}^{}\frac{1}{|{\cal C}_1^{j_1}|}Q_1\Biggl(\sum_{r_1=0}^{\tau_1-1}\nabla F_{j}(\mathbf{w}_{r_1,\ldots,r_{N}}^{j,t},\boldsymbol\xi_{r_1,\ldots,r_{N}}^{j,t})\Biggr)\Biggr)\nonumber\\&- \sum_{r_2=0}^{\tau_2-1}\sum_{j \in {\cal C}_1^{j_1}}^{}\frac{1}{|{\cal C}_1^{j_1}|}\sum_{r_1=0}^{\tau_1-1}\nabla F(\mathbf{w}_{r_1,\ldots,r_{N}}^{j,t})\Biggr\Vert^2\Biggr\}\Biggr] = \sum_{j_{2} \in {\cal C}_{3,i}}^{} \frac{|{\cal C}_{2}^{j_{2}}|^2}{|{\cal C}_{3,i}|^2}\nonumber\\&\sum_{r_4 = 0}^{t_{4}-1}\Biggl[q_3\mathbb{E}\Biggl\{\Biggl\Vert \sum_{r_3=0}^{\tau_3-1}\sum_{j_1\in {\cal C}_2^{j_2}}^{} \frac{|{\cal C}_1^{j_1}|}{|{\cal C}_2^{j_2}|}\sum_{r_2=0}^{\tau_2-1}\sum_{j \in {\cal C}_1^{j_1}}^{}\frac{1}{|{\cal C}_1^{j_1}|}\sum_{r_1=0}^{\tau_1-1}\nonumber\\&\nabla F(\mathbf{w}_{r_1,\ldots,r_{N}}^{j,t})\Biggr\Vert^2 \Biggr\}+ (1+q_3)\sum_{r_3=0}^{\tau_3-1}\sum_{j_1\in {\cal C}_2^{j_2}}^{} \frac{|{\cal C}_1^{j_1}|^2}{|{\cal C}_2^{j_2}|^2}\mathbb{E}\Biggl\{\Biggl\Vert Q_2\Biggl(\nonumber\\&\sum_{r_2=0}^{\tau_2-1}\sum_{j \in {\cal C}_1^{j_1}}^{}\frac{1}{|{\cal C}_1^{j_1}|}Q_1\Biggl(\sum_{r_1=0}^{\tau_1-1}\nabla F_{j}(\mathbf{w}_{r_1,\ldots,r_{N}}^{j,t},\boldsymbol\xi_{r_1,\ldots,r_{N}}^{j,t})\Biggr)\Biggr)\nonumber\\&- \sum_{r_2=0}^{\tau_2-1}\sum_{j \in {\cal C}_1^{j_1}}^{}\frac{1}{|{\cal C}_1^{j_1}|}\sum_{r_1=0}^{\tau_1-1}\nabla F(\mathbf{w}_{r_1,\ldots,r_{N}}^{j,t})\Biggr\Vert^2\Biggr\}\Biggr]= \sum_{j_{2} \in {\cal C}_{3,i}}^{} \frac{|{\cal C}_{2}^{j_{2}}|^2}{|{\cal C}_{3,i}|^2}\nonumber\\&\sum_{r_4 = 0}^{t_{4}-1}\Biggl[q_3\tau_1\tau_2\tau_3 \sum_{j_1\in {\cal C}_2^{j_2}}^{} \frac{1}{|{\cal C}_2^{j_2}|}\sum_{j \in {\cal C}_1^{j_1}}^{}\sum_{r_3=0}^{\tau_3-1}\sum_{r_2=0}^{\tau_2-1}\sum_{r_1=0}^{\tau_1-1}\mathbb{E}\Biggl\{\Biggl\Vert\nonumber\\&\nabla F(\mathbf{w}_{r_1,\ldots,r_{N}}^{j,t})\Biggr\Vert^2 \Biggr\}+ (1+q_3)\sum_{r_3=0}^{\tau_3-1}\sum_{j_1\in {\cal C}_2^{j_2}}^{} \frac{|{\cal C}_1^{j_1}|^2}{|{\cal C}_2^{j_2}|^2}\Biggl((1+q_2)\nonumber\\&(1+q_1)\tau_1\sigma^2\frac{1}{|{\cal C}_{1}^{j_1}|} \tau_2 + \frac{1}{|{\cal C}_{1}^{j_1}|}\left(q_2 \tau_1 \tau_2+ q_1 (1+q_2)\tau_1 \frac{1}{|{\cal C}_{1}^{j_1}|}\right)\nonumber\\&\sum_{j \in {\cal C}_{1}^{j_1}}^{}  \sum_{r_2=0}^{\tau_2-1}\sum_{r_1=0}^{\tau_1-1}\mathbb{E}\Biggl\{\Biggl\Vert\nabla F(\mathbf{w}_{r_1,\ldots,t_{N}}^{j,t})\Biggr\Vert^2\Biggr\}\Biggr)\Biggr] = \frac{1}{|{\cal C}_{3,i}|^2}\sum_{j_{2} \in {\cal C}_{3,i}}^{}  \nonumber\\&\sum_{j_1\in {\cal C}_2^{j_2}}^{} |{\cal C}_{2}^{j_{2}}|\biggl(q_3 \tau_1\tau_2 \tau_3 + \frac{{|{\cal C}_1^{j_1}|}}{|{\cal C}_{2}^{j_{2}}|}(1+q_3)\biggl(q_2 \tau_1 \tau_2+ \frac{1}{|{\cal C}_{1}^{j_1}|}q_1\nonumber\\& (1+q_2)\tau_1 \biggr)\biggr)\sum_{j \in {\cal C}_{1}^{j_1}}^{} \sum_{r_4=0}^{t_4-1}\sum_{r_3=0}^{\tau_3-1}\sum_{r_2=0}^{\tau_2-1}\sum_{r_1=0}^{\tau_1-1}\mathbb{E}\Biggl\{\Biggl\Vert\nabla F(\mathbf{w}_{r_1,\ldots,t_{N}}^{j,t})\Biggr\Vert^2\nonumber\\&\Biggr\}+ (1+q_3)(1+q_2)(1+q_1)t_4\tau_3\tau_2\tau_1\sigma^2  \frac{1}{|{\cal C}_{3,i}|}.
	\end{align}
	Following a similar approach with the steps given in~\eqref{4th} and ~\eqref{5th}, and leveraging their structural similarity for extension, we can bound \underline{the last term of the right side of~\eqref{difference}} as
	\begin{align}
		\label{Nth}
		&\mathbb{E}\Biggl\{\Biggl\Vert \sum_{j_{N-2} \in {\cal C}_{N-1,i}}^{} \frac{|{\cal C}_{N-2}^{j_{N-2}}|}{|{\cal C}_{N-1,i}|}\sum_{r_N = 0}^{t_{N}-1}Q_{N-1}\Biggl(\cdots \sum_{r_3=0}^{\tau_3-1}\sum_{j_1\in {\cal C}_2^{j_2}}^{} \frac{|{\cal C}_1^{j_1}|}{|{\cal C}_2^{j_2}|}\nonumber\\&Q_2\Biggl(\sum_{r_2=0}^{\tau_2-1}\sum_{j \in {\cal C}_1^{j_1}}^{}\frac{1}{|{\cal C}_1^{j_1}|}Q_1\Biggl(\sum_{r_1=0}^{\tau_1-1}\nabla F_{j}(\mathbf{w}_{r_1,\ldots,r_{N}}^{j,t},\boldsymbol\xi_{r_1,\ldots,r_{N}}^{j,t})\Biggr)\Biggr)\Biggr) \nonumber\\&-\cdots \sum_{r_3=0}^{\tau_3-1}\sum_{j_1\in {\cal C}_2^{j_2}}^{} \frac{|{\cal C}_1^{j_1}|}{|{\cal C}_2^{j_2}|}\sum_{r_2=0}^{\tau_2-1}\sum_{j \in {\cal C}_1^{j_1}}^{}\frac{1}{|{\cal C}_1^{j_1}|}\sum_{r_1=0}^{\tau_1-1}\nabla F(\mathbf{w}_{r_1,\ldots,r_{N}}^{j,t})\Biggr\Vert^2\nonumber\\
		&\Biggr\}\leq \frac{1}{|{\cal C}_{N-1,i}|^2}\sum_{j_{N-2} \in {\cal C}_{N-1,i}}^{}\cdots \sum_{j_1\in {\cal C}_2^{j_2}}^{} |{\cal C}_{N-2}^{j_{N-2}}|\biggl(q_{N-1}\nonumber\\
		& \tau_1\cdots \tau_{N-1} + \frac{{|{\cal C}_{N-3}^{j_{N-3}}|}}{|{\cal C}_{N-2}^{j_{N-2}}|}(1+q_{N-1})\biggl(q_{N-2} \tau_1 \cdots \tau_{N-2}+\nonumber\\& \frac{{|{\cal C}_{N-4}^{j_{N-4}}|}}{|{\cal C}_{N-3}^{j_{N-3}}|}\biggl(\cdots\biggr) \biggr)\biggr)\sum_{j \in {\cal C}_{1}^{j_1}}^{} \sum_{r_N=0}^{t_N-1}\sum_{r_{N-1}=0}^{\tau_{N-1}-1}\cdots\sum_{r_1=0}^{\tau_1-1}\mathbb{E}\Biggl\{\Biggl\Vert\nonumber\\&\nabla F(\mathbf{w}_{r_1,\ldots,t_{N}}^{j,t})\Biggr\Vert^2\Biggr\}+ (1+q_{N-1})\cdots (1+q_1)t_{N}\nonumber\\&\tau_{N-1}\cdots\tau_1\sigma^2  \frac{1}{|{\cal C}_{{N-1},i}|}.
	\end{align}
	Now, replacing~\eqref{purenorm}-\eqref{Nth} in~\eqref{difference} and then replacing the result in~\eqref{diffexp},~\eqref{main1} as \underline{the first term of~\eqref{lipexpp}} is obtained as
	\begin{align}
		\label{result1}
		&-\frac{\mu}{N_\text{tot}}\sum_{i}^{} \sum_{t_N = 0}^{\tau_{N}-1}\cdots \sum_{t_3=0}^{\tau_3-1}\sum_{t_2=0}^{\tau_2-1}\sum_{t_1=0}^{\tau_1-1}\mathbb{E}\Bigl\{\nabla F( \mathbf{w}_{t})^\top\nonumber\\
		&\nabla F(\mathbf{w}_{t_1,\ldots,t_{N}}^{i,t})\Bigr\} = -\frac{\mu}{2}\tau_N\cdots\tau_1 \mathbb{E}\left\{\|\nabla F( {\mathbf{w}}_{t})\|^2\right\}  -\frac{\mu}{2N_\text{tot}}\nonumber\\
		&\sum_{i}^{} \sum_{t_N = 0}^{\tau_{N}-1}\cdots \sum_{t_3=0}^{\tau_3-1}\sum_{t_2=0}^{\tau_2-1}\sum_{t_1=0}^{\tau_1-1}\mathbb{E}\left\{\|\nabla F(\mathbf{w}_{t_1,\ldots,t_{N}}^{i,t})\|^2\right\} +\frac{\mu}{2N_\text{tot}}\nonumber\\&\sum_{i}^{} \sum_{t_N = 0}^{\tau_{N}-1}\cdots \sum_{t_3=0}^{\tau_3-1}\sum_{t_2=0}^{\tau_2-1}\sum_{t_1=0}^{\tau_1-1} \nonumber\\&\mathbb{E}\left\{\|\nabla F( {\mathbf{w}}_{t})- \nabla F(\mathbf{w}_{t_1,\ldots,t_{N}}^{i,t})\|^2\right\},
	\end{align}
	where the last term is the summation of the following terms:
	\begin{align}
		&\frac{L^2\mu^3}{2N_\text{tot}}\sum_{i}^{} \sum_{t_N = 0}^{\tau_{N}-1}\cdots \sum_{t_3=0}^{\tau_3-1}\sum_{t_2=0}^{\tau_2-1}\sum_{t_1=0}^{\tau_1-1}\Biggl[t_1\sum_{r_1=0}^{t_1-1}\mathbb{E}\Biggl\{\Biggl\Vert\nabla F(\mathbf{w}_{r_1,\ldots,t_{N}}^{i,t})\nonumber\\&\Biggr\Vert^2\Biggr\}+ t_2 \tau_1 \frac{1}{|{\cal C}_{1,i}|} \sum_{j \in {\cal C}_{1,i}}^{}\sum_{r_2=0}^{t_2-1}\sum_{r_1=0}^{\tau_1-1}\mathbb{E}\Biggl\{\Biggl\Vert\nabla F(\mathbf{w}_{r_1, r_2,\ldots,t_{N}}^{j,t})\Biggr\Vert^2\Biggr\} \nonumber\\&+ t_3 \tau_2 \tau_1 \frac{1}{|{\cal C}_{2,i}|}\sum_{j_1\in {\cal C}_{2,i}}^{}\sum_{j \in {\cal C}_{1}^{j_1}}^{} \sum_{r_3=0}^{t_3-1}\sum_{r_2=0}^{\tau_2-1}\sum_{r_1=0}^{\tau_1-1}\mathbb{E}\Biggl\{\Biggl\Vert\nonumber\\&\nabla F(\mathbf{w}_{r_1,r_2,r_3,\ldots,t_{N}}^{j,t})\Biggr\Vert^2\Biggr\}+\cdots+t_N \tau_{N-1} \cdots \tau_1 \frac{1}{|{\cal C}_{N-1,i}|} \nonumber
	\end{align}
	\begin{align}
		&\sum_{j_{N-2} \in {\cal C}_{N-1,i}}^{} \cdots \sum_{j \in {\cal C}_1^{j_1}}^{}\sum_{r_N = 0}^{t_{N}-1}\cdots  \sum_{r_2=0}^{\tau_2-1}\sum_{r_1=0}^{\tau_1-1}\mathbb{E}\Biggl\{\Biggl\Vert\nabla F(\mathbf{w}_{r_1,\ldots,r_{N}}^{j,t})\Biggr\Vert^2\nonumber\\&\Biggr\}\Biggr] \leq \frac{L^2\mu^3}{2N_\text{tot}}\Biggl[\sum_{t_1=0}^{\tau_1-1}t_1+\sum_{t_2=0}^{\tau_2-1}\sum_{t_1=0}^{\tau_1-1}t_2\tau_1+\sum_{t_3 = 0}^{\tau_3-1}\sum_{t_2=0}^{\tau_2-1}\sum_{t_1=0}^{\tau_1-1}t_3\tau_2\tau_1\nonumber\\&+\cdots+\sum_{t_N = 0}^{\tau_N-1}\cdots\sum_{t_1=0}^{\tau_1-1}t_N\tau_{N-1}\cdots\tau_1\Biggr]\sum_{i}^{} \sum_{t_N = 0}^{\tau_{N}-1}\cdots \sum_{t_1=0}^{\tau_1-1} \mathbb{E}\Biggl\{\nonumber\\&\Biggl\Vert\nabla F(\mathbf{w}_{t_1,\ldots,t_{N}}^{i,t})\Biggr\Vert^2\Biggr\}= \frac{L^2\mu^3}{2N_\text{tot}}\Biggl[\frac{\tau_1(\tau_1-1)}{2}+\tau_1^2 \frac{\tau_2(\tau_2-1)}{2}+\nonumber\\&\tau_1^2\tau_2^2\frac{\tau_3 (\tau_3-1)}{2}+\cdots+\tau_1^2\cdots\tau_{N-1}^2 \frac{\tau_N (\tau_N-1)}{2}\Biggr]\sum_{i}^{} \sum_{t_N = 0}^{\tau_{N}-1}\nonumber\\&\cdots \sum_{t_1=0}^{\tau_1-1} \mathbb{E}\Biggl\{\Biggl\Vert\nabla F(\mathbf{w}_{t_1,\ldots,t_{N}}^{i,t})\Biggr\Vert^2\Biggr\},
	\end{align}
	and
	
	\begin{align}
		&\frac{L^2\mu^3}{2N_\text{tot}}\sum_{i}^{} \sum_{t_N = 0}^{\tau_{N}-1}\cdots \sum_{t_1=0}^{\tau_1-1}t_1 \sigma^2 = \frac{L^2\mu^3}{2}\sigma^2 \tau_N \cdots \tau_2 \frac{\tau_1(\tau_1-1)}{2},
	\end{align}
	and
	\begin{align}
		&\frac{L^2\mu^3}{2N_\text{tot}}\sum_{i}^{} \sum_{t_N = 0}^{\tau_{N}-1}\cdots \sum_{t_1=0}^{\tau_1-1}\Biggl[(1+q_1)  t_2 \tau_1 \sigma^2 \frac{1}{|{\cal C}_{1,i}|} + \frac{1}{|{\cal C}_{1,i}|^2} q_1\tau_1 \nonumber\\&\sum_{j \in {\cal C}_{1,i}}^{}\sum_{r_2=0}^{t_2-1}\sum_{r_1=0}^{\tau_1-1}\mathbb{E}\Biggl\{\Biggl\Vert\nabla F(\mathbf{w}_{r_1, r_2,\ldots,t_{N}}^{j,t})\Biggr\Vert^2\Biggr\}\Biggr]\leq \frac{L^2\mu^3}{2N_\text{tot}}\sum_{i}^{} \sum_{t_N = 0}^{\tau_{N}-1}\nonumber\\
		&\cdots \sum_{t_1=0}^{\tau_1-1}(1+q_1)  t_2 \tau_1 \sigma^2\frac{1}{|{\cal C}_{1,i}|} + \frac{L^2\mu^3}{2N_\text{tot}}\sum_{i}^{}\frac{1}{|{\cal C}_{1,i}|} \sum_{j \in {\cal C}_{1,i}}^{}\sum_{t_N = 0}^{\tau_{N}-1}\nonumber\\
		&\cdots \sum_{t_1=0}^{\tau_1-1}  q_1\tau_1 \sum_{r_2=0}^{t_2-1}\sum_{r_1=0}^{\tau_1-1}\mathbb{E}\Biggl\{\Biggl\Vert\nabla F(\mathbf{w}_{r_1, r_2,\ldots,t_{N}}^{j,t})\Biggr\Vert^2\Biggr\} = \frac{L^2\mu^3}{2N_\text{tot}} \nonumber\\&(1+q_1)\tau_N \cdots \tau_3 \frac{\tau_2(\tau_2-1)}{2} \tau_1^2 \sigma^2 \sum_{i} \frac{1}{|{\cal C}_{1,i}|}+ \frac{L^2\mu^3}{2N_\text{tot}}\sum_{i}^{}\frac{1}{|{\cal C}_{1,i}|} \nonumber\\&\sum_{j \in {\cal C}_{1,i}}^{}\sum_{t_N = 0}^{\tau_{N}-1}\cdots \sum_{t_1=0}^{\tau_1-1} \mathbb{E}\Biggl\{\Biggl\Vert\nabla F(\mathbf{w}_{t_1, t_2,\ldots,t_{N}}^{j,t})\Biggr\Vert^2\Biggr\}  \sum_{r_2=0}^{\tau_2-1}\sum_{r_1=0}^{\tau_1-1}q_1\tau_1 \nonumber\\
		&= \frac{L^2\mu^3}{2N_\text{tot}} (1+q_1)\tau_N \cdots \tau_3 \frac{\tau_2(\tau_2-1)}{2} \tau_1^2 \sigma^2  C_1+\frac{L^2\mu^3}{2N_\text{tot}}q_1 \tau_2 \tau_1^2\nonumber\\
		&\sum_{i}^{}\sum_{t_N = 0}^{\tau_{N}-1}\cdots \sum_{t_1=0}^{\tau_1-1} \mathbb{E}\Biggl\{\Biggl\Vert\nabla F(\mathbf{w}_{t_1, t_2,\ldots,t_{N}}^{i,t})\Biggr\Vert^2\Biggr\},
	\end{align}
	and
	\begin{align}
		&\frac{L^2\mu^3}{2N_\text{tot}}\sum_{i}^{} \sum_{t_N = 0}^{\tau_{N}-1}\cdots \sum_{t_1=0}^{\tau_1-1}\Biggl[(1+q_2)(1+q_1)t_3\tau_2 \tau_1\sigma^2\frac{1}{|{\cal C}_{2,i}|}  +\nonumber\\& \frac{1}{|{\cal C}_{2,i}|^2}\sum_{j_1\in {\cal C}_{2,i}}^{} {|{\cal C}_{1}^{j_1}|}\left(q_2 \tau_1 \tau_2+\frac{1}{|{\cal C}_{1}^{j_1}|} q_1 (1+q_2)\tau_1 \right)\sum_{j \in {\cal C}_{1}^{j_1}}^{}\nonumber\\&\sum_{r_3=0}^{t_3-1}  \sum_{r_2=0}^{\tau_2-1}\sum_{r_1=0}^{\tau_1-1}\mathbb{E}\Biggl\{\Biggl\Vert\nabla F(\mathbf{w}_{r_1,\ldots,t_{N}}^{j,t})\Biggr\Vert^2\Biggr\}\Biggr]=\frac{L^2\mu^3}{2N_\text{tot}}(1+q_2)\nonumber
	\end{align}
	\begin{align}
		&(1+q_1)\tau_N \cdots \tau_4 \frac{\tau_3(\tau_3-1)}{2} \tau_2^2 \tau_1^2 \sigma^2\sum_{i}^{}\frac{1}{|{\cal C}_{2,i}|} +\frac{L^2\mu^3}{2N_\text{tot}}\sum_{i}^{}\nonumber\\
		&\frac{1}{|{\cal C}_{2,i}|^2}\sum_{j_1\in {\cal C}_{2,i}}^{} {|{\cal C}_{1}^{j_1}|}\left(q_2 \tau_1 \tau_2+\frac{1}{|{\cal C}_{1}^{j_1}|} q_1 (1+q_2)\tau_1 \right)\sum_{j \in {\cal C}_{1}^{j_1}}^{}\sum_{t_N = 0}^{\tau_{N}-1}\nonumber\\
		&\cdots \sum_{t_1=0}^{\tau_1-1}\mathbb{E}\Biggl\{\Biggl\Vert\nabla F(\mathbf{w}_{t_1,\ldots,t_{N}}^{j,t})\Biggr\Vert^2\Biggr\} \times \sum_{t_3 = 0}^{\tau_{3}-1}\sum_{t_2 = 0}^{\tau_{2}-1}\sum_{t_1 = 0}^{\tau_{1}-1}1  \leq \frac{L^2\mu^3}{2N_\text{tot}}\nonumber\\&(1+q_2)(1+q_1)\tau_N \cdots \tau_4 \frac{\tau_3(\tau_3-1)}{2} \tau_2^2 \tau_1^2 \sigma^2C_2 +\frac{L^2\mu^3}{2N_\text{tot}}\tau_3 \tau_2 \tau_1\nonumber\\&\sum_{i}^{}\frac{1}{|{\cal C}_{2,i}|}\sum_{j_1\in {\cal C}_{2,i}}^{} {|{\cal C}_{1}^{j_1}|}\left(q_2 \tau_1 \tau_2+\frac{1}{|{\cal C}_{1}^{j_1}|} q_1 (1+q_2)\tau_1 \right)\sum_{j \in {\cal C}_{1}^{j_1}}^{}\nonumber\\&\sum_{t_N = 0}^{\tau_{N}-1}\cdots \sum_{t_1=0}^{\tau_1-1}\mathbb{E}\Biggl\{\Biggl\Vert\nabla F(\mathbf{w}_{t_1,\ldots,t_{N}}^{j,t})\Biggr\Vert^2\Biggr\} =\frac{L^2\mu^3}{2N_\text{tot}}(1+q_2)\nonumber\\&(1+q_1)\tau_N \cdots \tau_4 \frac{\tau_3(\tau_3-1)}{2} \tau_2^2 \tau_1^2 \sigma^2C_2 +\frac{L^2\mu^3}{2N_\text{tot}}\tau_3 \tau_2 \tau_1\sum_{i}^{}\nonumber\\&{|{\cal C}_{1,i}|}\left(q_2 \tau_1 \tau_2+\frac{1}{|{\cal C}_{1,i}|} q_1 (1+q_2)\tau_1 \right)\sum_{t_N = 0}^{\tau_{N}-1}\cdots \sum_{t_1=0}^{\tau_1-1}\mathbb{E}\Biggl\{\Biggl\Vert\nonumber\\&\nabla F(\mathbf{w}_{t_1,\ldots,t_{N}}^{i,t})\Biggr\Vert^2\Biggr\}=\frac{L^2\mu^3}{2N_\text{tot}}(1+q_2)(1+q_1)\tau_N \cdots \tau_4 \nonumber\\&\frac{\tau_3(\tau_3-1)}{2} \tau_2^2 \tau_1^2 \sigma^2C_2 +\frac{L^2\mu^3}{2N_\text{tot}}\tau_3 \tau_2 \tau_1 \max_{} \Biggl\{{|{\cal C}_{1}^{i_1}|}\biggl(q_2 \tau_1 \tau_2+\nonumber\\
		&\frac{1}{|{\cal C}_{1}^{i_1}|} q_1 (1+q_2)\tau_1 \biggr)\Biggr\}\sum_{i}^{}\sum_{t_N = 0}^{\tau_{N}-1}\cdots \sum_{t_1=0}^{\tau_1-1}\nonumber\\
		&\mathbb{E}\Biggl\{\Biggl\Vert\nabla F(\mathbf{w}_{t_1,\ldots,t_{N}}^{i,t})\Biggr\Vert^2\Biggr\},
	\end{align}
	where, in the last step, we used the bound $\sum_{i}^{} u_i v_i \leq \max_{}u_i \cdot\sum_i v_i$. The remaining portion of the summation in the last term of~\eqref{result1} is then obtained as
	\begin{align}
		&\frac{L^2\mu^3}{2N_\text{tot}}\sum_{i}^{} \sum_{t_N = 0}^{\tau_{N}-1}\cdots \sum_{t_1=0}^{\tau_1-1}\Biggl[(1+q_{N-1})\cdots (1+q_1)t_{N}\tau_{N-1}\cdots\tau_1\nonumber\\&\sigma^2  \frac{1}{|{\cal C}_{{N-1},i}|}+\frac{1}{|{\cal C}_{N-1,i}|^2}\sum_{j_{N-2} \in {\cal C}_{N-1,i}}^{}\cdots \sum_{j_1\in {\cal C}_2^{j_2}}^{} |{\cal C}_{N-2}^{j_{N-2}}|\biggl(q_{N-1}\nonumber\\& \tau_1\cdots \tau_{N-1} + \frac{{|{\cal C}_{N-3}^{j_{N-3}}|}}{|{\cal C}_{N-2}^{j_{N-2}}|}(1+q_{N-1})\biggl(q_{N-2} \tau_1 \cdots \tau_{N-2}+\nonumber\\& \frac{{|{\cal C}_{N-4}^{j_{N-4}}|}}{|{\cal C}_{N-3}^{j_{N-3}}|}\biggl(\cdots\biggr) \biggr)\biggr)\sum_{j \in {\cal C}_{1}^{j_1}}^{} \sum_{r_N=0}^{t_N-1}\sum_{r_{N-1}=0}^{\tau_{N-1}-1}\cdots\sum_{r_1=0}^{\tau_1-1}\nonumber\\&\mathbb{E}\Biggl\{\Biggl\Vert\nabla F(\mathbf{w}_{r_1,\ldots,t_{N}}^{j,t})\Biggr\Vert^2\Biggr\}\Biggr] \leq  \frac{L^2\mu^3}{2N_\text{tot}}(1+q_{N-1})\cdots(1+q_1)\nonumber\\&\frac{\tau_N(\tau_N-1)}{2}\tau_{N-1}^2 \cdots \tau_1^2 \sigma^2C_{N-1}+\frac{L^2\mu^3}{2N_\text{tot}}\tau_{N} \cdots \tau_1 \max_{} \Biggl\{ \nonumber\\
		&|{\cal C}_{N-2}^{i_{N-2}}|\biggl(q_{N-1} \tau_1\cdots \tau_{N-1} + \frac{{|{\cal C}_{N-3}^{i_{N-3}}|}}{|{\cal C}_{N-2}^{i_{N-2}}|}(1+q_{N-1})\biggl(q_{N-2} \tau_1 \nonumber
	\end{align}
	\begin{align}
		&\cdots \tau_{N-2}+ \frac{{|{\cal C}_{N-4}^{i_{N-4}}|}}{|{\cal C}_{N-3}^{i_{N-3}}|}\biggl(\cdots\biggr) \biggr)\biggr)\Biggr\}\sum_{i}^{}\sum_{t_N = 0}^{\tau_{N}-1}\cdots \sum_{t_1=0}^{\tau_1-1}\mathbb{E}\Biggl\{\Biggl\Vert\nonumber\\&\nabla F(\mathbf{w}_{t_1,\ldots,t_{N}}^{i,t})\Biggr\Vert^2\Biggr\}.
	\end{align}
	\textbf{Second Term of~\eqref{lipexpp}:} From~\eqref{purenorm} and~\eqref{Nth}, we can bound this term as
	\begin{align}
		\label{result2}
		&\mathbb{E}\left\{\|{\mathbf{w}}_{t+1} - {\mathbf{w}}_{t}\|^2\right\} = \mu^2 \mathbb{E}\Biggl\{\Biggl\Vert \sum_{i_{N-1}}^{} \frac{|{\cal C}_{N-1}^{i_{N-1}}|}{N_\text{tot}}Q_N\Biggl(\sum_{t_N = 0}^{\tau_{N}-1}\nonumber\\&\sum_{i_{N-2} \in {\cal C}_{N-1}^{i_{N-1}}}^{} \frac{|{\cal C}_{N-2}^{i_{N-2}}|}{|{\cal C}_{N-1}^{i_{N-1}}|}Q_{N-1}\Biggl(\cdots \sum_{t_3=0}^{\tau_3-1}\sum_{i_1\in {\cal C}_2^{i_2}}^{} \frac{|{\cal C}_1^{i_1}|}{|{\cal C}_2^{i_2}|}Q_2\Biggl(\sum_{t_2=0}^{\tau_2-1}\sum_{i \in {\cal C}_1^{i_1}}^{}\nonumber\\&\frac{1}{|{\cal C}_1^{i_1}|}Q_1\Biggl(\sum_{t_1=0}^{\tau_1-1}\nabla F_{i}(\mathbf{w}_{t_1,\ldots,t_{N}}^{i,t},\boldsymbol\xi_{t_1,\ldots,t_{N}}^{i,t})\Biggr)\Biggr)\Biggr)\Biggr)\Biggr\Vert^2\Biggr\}  =  \frac{\mu^2}{N_\text{tot}^2}\nonumber\\&\mathbb{E}\Biggl\{\Biggl\Vert\sum_{i_{N-1}}\sum_{i_{N-2} \in {\cal C}_{N-1}^{i_{N-1}}}^{} \cdots \sum_{i \in {\cal C}_1^{i_1}}^{}\sum_{t_N = 0}^{\tau_{N}-1}\cdots  \sum_{t_2=0}^{\tau_2-1}\sum_{t_1=0}^{\tau_1-1}\nonumber\\&\nabla F(\mathbf{w}_{t_1,\ldots,t_{N}}^{i,t})\Biggr\Vert^2\Biggr\}+\mu^2\mathbb{E}\Biggl\{\Biggl\Vert \sum_{i_{N-1}}^{} \frac{|{\cal C}_{N-1}^{i_{N-1}}|}{N_\text{tot}}Q_N\Biggl(\cdots \sum_{t_3=0}^{\tau_3-1}\sum_{i_1\in {\cal C}_2^{i_2}}^{} \nonumber\\&\frac{|{\cal C}_1^{i_1}|}{|{\cal C}_2^{i_2}|}Q_2\Biggl(\sum_{t_2=0}^{\tau_2-1}\sum_{i \in {\cal C}_1^{i_1}}^{}\frac{1}{|{\cal C}_1^{i_1}|}Q_1\Biggl(\sum_{t_1=0}^{\tau_1-1}\nabla F_{i}(\mathbf{w}_{t_1,\ldots,t_{N}}^{i,t},\boldsymbol\xi_{t_1,\ldots,t_{N}}^{i,t})\Biggr)\nonumber\\&\Biggr)\Biggr) -\cdots \sum_{t_3=0}^{\tau_3-1}\sum_{i_1\in {\cal C}_2^{i_2}}^{} \frac{|{\cal C}_1^{i_1}|}{|{\cal C}_2^{i_2}|}\sum_{t_2=0}^{\tau_2-1}\sum_{i \in {\cal C}_1^{i_1}}^{}\frac{1}{|{\cal C}_1^{i_1}|}\sum_{t_1=0}^{\tau_1-1}\nabla F(\mathbf{w}_{t_1,\ldots,t_{N}}^{i,t})\nonumber\\&\Biggr\Vert^2\Biggr\} \leq \frac{\mu^2}{N_\text{tot}} \tau_N \cdots \tau_1 \sum_{i_{N-1}}\sum_{i_{N-2} \in {\cal C}_{N-1}^{i_{N-1}}}^{} \cdots \sum_{i \in {\cal C}_1^{i_1}}^{}\sum_{t_N = 0}^{\tau_{N}-1}\cdots  \sum_{t_2=0}^{\tau_2-1}\nonumber\\&\sum_{t_1=0}^{\tau_1-1}\mathbb{E}\Biggl\{\Biggl\Vert\nabla F(\mathbf{w}_{t_1,\ldots,t_{N}}^{i,t})\Biggr\Vert^2\Biggr\}+\frac{\mu^2}{N_\text{tot}^2}\sum_{i_{N-1}}^{}\cdots \sum_{i_1\in {\cal C}_2^{i_2}}^{} |{\cal C}_{N-1}^{i_{N-1}}|\biggl(\nonumber\\&q_{N} \tau_1\cdots \tau_{N} + \frac{{|{\cal C}_{N-2}^{i_{N-2}}|}}{|{\cal C}_{N-1}^{i_{N-1}}|}(1+q_N)\biggl(q_{N-1} \tau_1 \cdots \tau_{N-1}+ \frac{{|{\cal C}_{N-3}^{i_{N-3}}|}}{|{\cal C}_{N-2}^{i_{N-2}}|}\nonumber\\&\biggl(\cdots\biggr) \biggr)\biggr)\sum_{i \in {\cal C}_{1}^{i_1}}^{} \sum_{t_N=0}^{\tau_N-1}\sum_{t_{N-1}=0}^{\tau_{N-1}-1}\cdots\sum_{t_1=0}^{\tau_1-1}\mathbb{E}\Biggl\{\Biggl\Vert\nabla F(\mathbf{w}_{t_1,\ldots,t_{N}}^{i,t})\Biggr\Vert^2\Biggr\}\nonumber\\&+ \mu^2 (1+q_{N})\cdots (1+q_1)\tau_{N}\tau_{N-1}\cdots\tau_1\sigma^2  \frac{1}{N_\text{tot}}.
	\end{align}
	\textbf{Final Bound on~\eqref{lipexpp}:} We now have both terms in~\eqref{lipexpp}, derived from~\eqref{result1} and~\eqref{result2}, and thus obtain
	\begin{align}
		\label{lipexp}
		&\mathbb{E}\left\{F({\mathbf{w}}_{t+1})-F({\mathbf{w}}_{t})\right\} \leq -\frac{\mu}{2}\tau_N\cdots\tau_1 \mathbb{E}\left\{\|\nabla F( {\mathbf{w}}_{t})\|^2\right\}  -\nonumber\\&\frac{\mu}{2N_\text{tot}}\sum_{i}^{} \sum_{t_N = 0}^{\tau_{N}-1}\cdots \sum_{t_3=0}^{\tau_3-1}\sum_{t_2=0}^{\tau_2-1}\sum_{t_1=0}^{\tau_1-1}\mathbb{E}\left\{\|\nabla F(\mathbf{w}_{t_1,\ldots,t_{N}}^{i,t})\|^2\right\}+\nonumber\\
		&\frac{L^2\mu^3}{2N_\text{tot}}\Biggl[\frac{\tau_1(\tau_1-1)}{2}+\tau_1^2 \frac{\tau_2(\tau_2-1)}{2}+\tau_1^2\tau_2^2\frac{\tau_3 (\tau_3-1)}{2}+\cdots\nonumber\\
		&+\tau_1^2\cdots\tau_{N-1}^2 \frac{\tau_N (\tau_N-1)}{2}\Biggr]\sum_{i}^{} \sum_{t_N = 0}^{\tau_{N}-1}\cdots \sum_{t_1=0}^{\tau_1-1} \mathbb{E}\Biggl\{\Biggl\Vert\nonumber
	\end{align}
	\begin{align}
		&\nabla F(\mathbf{w}_{t_1,\ldots,t_{N}}^{i,t})\Biggr\Vert^2\Biggr\}+\frac{L^2\mu^3}{2}\sigma^2 \tau_N \cdots \tau_2 \frac{\tau_1(\tau_1-1)}{2}+\frac{L^2\mu^3}{2N_\text{tot}} \nonumber\\&(1+q_1)\tau_N \cdots \tau_3 \frac{\tau_2(\tau_2-1)}{2} \tau_1^2 \sigma^2  C_1+\frac{L^2\mu^3}{2N_\text{tot}}q_1 \tau_2 \tau_1^2\sum_{i}^{}\sum_{t_N = 0}^{\tau_{N}-1}\nonumber\\&\cdots \sum_{t_1=0}^{\tau_1-1} \mathbb{E}\Biggl\{\Biggl\Vert\nabla F(\mathbf{w}_{t_1, t_2,\ldots,t_{N}}^{i,t})\Biggr\Vert^2\Biggr\}+\frac{L^2\mu^3}{2N_\text{tot}}(1+q_2)(1+q_1)\nonumber\\&\tau_N \cdots \tau_4 \frac{\tau_3(\tau_3-1)}{2} \tau_2^2 \tau_1^2 \sigma^2C_2 +\frac{L^2\mu^3}{2N_\text{tot}}\tau_3 \tau_2 \tau_1 \max_{} \Biggl\{{|{\cal C}_{1}^{i_1}|}\biggl(q_2 \nonumber\\&\tau_1 \tau_2+\frac{1}{|{\cal C}_{1}^{i_1}|} q_1 (1+q_2)\tau_1 \biggr)\Biggr\}\sum_{i}^{}\sum_{t_N = 0}^{\tau_{N}-1}\cdots \sum_{t_1=0}^{\tau_1-1}\mathbb{E}\Biggl\{\Biggl\Vert\nonumber\\
		&\nabla F(\mathbf{w}_{t_1,\ldots,t_{N}}^{i,t})\Biggr\Vert^2\Biggr\}+\cdots+\frac{L^2\mu^3}{2N_\text{tot}}(1+q_{N-1})\cdots(1+q_1)\nonumber\\
		&\frac{\tau_N(\tau_N-1)}{2}\tau_{N-1}^2 \cdots \tau_1^2 \sigma^2C_{N-1}+\frac{L^2\mu^3}{2N_\text{tot}}\tau_{N} \cdots \tau_1 \max_{} \Biggl\{ \nonumber\\
		&|{\cal C}_{N-2}^{i_{N-2}}|\biggl(q_{N-1} \tau_1\cdots \tau_{N-1} + \frac{{|{\cal C}_{N-3}^{i_{N-3}}|}}{|{\cal C}_{N-2}^{i_{N-2}}|}(1+q_{N-1})\biggl(q_{N-2} \tau_1 \nonumber\\
		&\cdots \tau_{N-2}+ \frac{{|{\cal C}_{N-4}^{i_{N-4}}|}}{|{\cal C}_{N-3}^{i_{N-3}}|}\biggl(\cdots\biggr) \biggr)\biggr)\Biggr\}\sum_{i}^{}\sum_{t_N = 0}^{\tau_{N}-1}\cdots \sum_{t_1=0}^{\tau_1-1}\mathbb{E}\Biggl\{\Biggl\Vert\nonumber\\
		&\nabla F(\mathbf{w}_{t_1,\ldots,t_{N}}^{i,t})\Biggr\Vert^2\Biggr\}+\frac{L}{2} \frac{\mu^2}{N_\text{tot}} \tau_N \cdots \tau_1 \sum_{i}\sum_{t_N = 0}^{\tau_{N}-1}\cdots  \sum_{t_2=0}^{\tau_2-1}\nonumber\\
		&\sum_{t_1=0}^{\tau_1-1}\mathbb{E}\Biggl\{\Biggl\Vert\nabla F(\mathbf{w}_{t_1,\ldots,t_{N}}^{i,t})\Biggr\Vert^2\Biggr\}+\frac{L}{2}\frac{\mu^2}{N_\text{tot}^2} \max_{} \Biggl\{|{\cal C}_{N-1}^{i_{N-1}}|\biggl(q_{N} \tau_1\nonumber\\
		&\cdots \tau_{N} + \frac{{|{\cal C}_{N-2}^{i_{N-2}}|}}{|{\cal C}_{N-1}^{i_{N-1}}|}(1+q_N)\biggl(q_{N-1} \tau_1 \cdots \tau_{N-1}+ \frac{{|{\cal C}_{N-3}^{i_{N-3}}|}}{|{\cal C}_{N-2}^{i_{N-2}}|}\nonumber\\&\biggl(\cdots\biggr) \biggr)\biggr)\Biggr\}\sum_{i}^{} \sum_{t_N=0}^{\tau_N-1}\sum_{t_{N-1}=0}^{\tau_{N-1}-1}\cdots\sum_{t_1=0}^{\tau_1-1}\mathbb{E}\Biggl\{\Biggl\Vert\nabla F(\mathbf{w}_{t_1,\ldots,t_{N}}^{i,t})\Biggr\Vert^2\Biggr\}\nonumber\\&+ \frac{L}{2}\mu^2(1+q_{N})\cdots (1+q_1)\tau_{N}\tau_{N-1}\cdots\tau_1\sigma^2  \frac{1}{N_\text{tot}},
	\end{align}
	which can be written as
	\begin{align}
		\label{minus_bound}
		&\mathbb{E}\left\{F({\mathbf{w}}_{t+1})-F({\mathbf{w}}_{t})\right\} \leq -\frac{\mu}{2}\tau_N\cdots\tau_1 \mathbb{E}\left\{\|\nabla F( {\mathbf{w}}_{t})\|^2\right\}+\nonumber\\&\frac{L^2\mu^3}{2}\sigma^2 \tau_N \cdots \tau_2 \frac{\tau_1(\tau_1-1)}{2}+\frac{L^2\mu^3}{2N_\text{tot}} (1+q_1)\tau_N \cdots \tau_3 \nonumber\\&\frac{\tau_2(\tau_2-1)}{2} \tau_1^2 \sigma^2  C_1+\frac{L^2\mu^3}{2N_\text{tot}}(1+q_2)(1+q_1)\tau_N \cdots \tau_4\nonumber\\& \frac{\tau_3(\tau_3-1)}{2} \tau_2^2 \tau_1^2 \sigma^2C_2+\cdots+\frac{L^2\mu^3}{2N_\text{tot}}(1+q_{N-1})\cdots(1+q_1)\nonumber\\&\frac{\tau_N(\tau_N-1)}{2}\tau_{N-1}^2 \cdots \tau_1^2 \sigma^2C_{N-1}+\frac{L}{2}\mu^2(1+q_{N})\cdots (1+q_1)\nonumber\\
		&\tau_{N}\tau_{N-1}\cdots\tau_1\sigma^2  \frac{1}{N_\text{tot}}+\sum_{i}^{} \sum_{t_N = 0}^{\tau_{N}-1}\cdots \sum_{t_3=0}^{\tau_3-1}\sum_{t_2=0}^{\tau_2-1}\sum_{t_1=0}^{\tau_1-1}\nonumber\\
		&\mathbb{E}\left\{\|\nabla F(\mathbf{w}_{t_1,\ldots,t_{N}}^{i,t})\|^2\right\}\times\Biggl[-\frac{\mu}{2N_\text{tot}}+\frac{L^2\mu^3}{2N_\text{tot}}\Biggl[\frac{\tau_1(\tau_1-1)}{2}+\nonumber
	\end{align}
	\begin{align}
		&\tau_1^2 \frac{\tau_2(\tau_2-1)}{2}+\tau_1^2\tau_2^2\frac{\tau_3 (\tau_3-1)}{2}+\cdots+\tau_1^2\cdots\tau_{N-1}^2 \nonumber\\
		&\frac{\tau_N (\tau_N-1)}{2}\Biggr]+\frac{L^2\mu^3}{2N_\text{tot}}q_1 \tau_2 \tau_1^2+\frac{L^2\mu^3}{2N_\text{tot}}\tau_3 \tau_2 \tau_1 \max_{} \Biggl\{{|{\cal C}_{1}^{i_1}|}\biggl(q_2\nonumber\\& \tau_1 \tau_2+\frac{1}{|{\cal C}_{1}^{i_1}|} q_1 (1+q_2)\tau_1 \biggr)\Biggr\}+\frac{L^2\mu^3}{2N_\text{tot}}\tau_{N} \cdots \tau_1 \max_{} \Biggl\{ |{\cal C}_{N-2}^{i_{N-2}}|\nonumber\\&\biggl(q_{N-1} \tau_1\cdots \tau_{N-1} + \frac{{|{\cal C}_{N-3}^{i_{N-3}}|}}{|{\cal C}_{N-2}^{i_{N-2}}|}(1+q_{N-1})\biggl(q_{N-2} \tau_1 \cdots \tau_{N-2}\nonumber\\&+ \frac{{|{\cal C}_{N-4}^{i_{N-4}}|}}{|{\cal C}_{N-3}^{i_{N-3}}|}\biggl(\cdots\biggr) \biggr)\biggr)\Biggr\}+\frac{L}{2} \frac{\mu^2}{N_\text{tot}} \tau_N \cdots \tau_1+\frac{L}{2}\frac{\mu^2}{N_\text{tot}^2} \max_{} \Biggl\{\nonumber\\&|{\cal C}_{N-1}^{i_{N-1}}|\biggl(q_{N} \tau_1\cdots \tau_{N} + \frac{{|{\cal C}_{N-2}^{i_{N-2}}|}}{|{\cal C}_{N-1}^{i_{N-1}}|}(1+q_N)\biggl(q_{N-1} \tau_1 \cdots \tau_{N-1}\nonumber\\
		&+ \frac{{|{\cal C}_{N-3}^{i_{N-3}}|}}{|{\cal C}_{N-2}^{i_{N-2}}|}\biggl(\cdots\biggr) \biggr)\biggr)\Biggr\}\Biggr].
	\end{align}
	Thus, if the following condition holds:
	\begin{align}
		&{\cal J} = -\frac{\mu}{2N_\text{tot}}+\frac{L^2\mu^3}{2N_\text{tot}}\Biggl[\frac{\tau_1(\tau_1-1)}{2}+\tau_1^2 \frac{\tau_2(\tau_2-1)}{2}+\tau_1^2\tau_2^2\nonumber\\&\frac{\tau_3 (\tau_3-1)}{2}+\cdots+\tau_1^2\cdots\tau_{N-1}^2 \frac{\tau_N (\tau_N-1)}{2}\Biggr]+\frac{L^2\mu^3}{2N_\text{tot}}q_1 \tau_2 \tau_1^2\nonumber\\&+\frac{L^2\mu^3}{2N_\text{tot}}\tau_3 \tau_2 \tau_1 \max_{} \Biggl\{{|{\cal C}_{1}^{i_1}|}\left(q_2 \tau_1 \tau_2+\frac{1}{|{\cal C}_{1}^{i_1}|} q_1 (1+q_2)\tau_1 \right)\Biggr\}\nonumber\\&+\frac{L^2\mu^3}{2N_\text{tot}}\tau_{N} \cdots \tau_1 \max_{} \Biggl\{ |{\cal C}_{N-2}^{i_{N-2}}|\biggl(q_{N-1} \tau_1\cdots \tau_{N-1} + \frac{{|{\cal C}_{N-3}^{i_{N-3}}|}}{|{\cal C}_{N-2}^{i_{N-2}}|}\nonumber\\&(1+q_{N-1})\biggl(q_{N-2} \tau_1 \cdots \tau_{N-2}+ \frac{{|{\cal C}_{N-4}^{i_{N-4}}|}}{|{\cal C}_{N-3}^{i_{N-3}}|}\biggl(\cdots\biggr) \biggr)\biggr)\Biggr\}+\frac{L}{2} \frac{\mu^2}{N_\text{tot}} \nonumber\\&\tau_N \cdots \tau_1+\frac{L}{2}\frac{\mu^2}{N_\text{tot}^2} \max_{} \Biggl\{|{\cal C}_{N-1}^{i_{N-1}}|\biggl(q_{N} \tau_1\cdots \tau_{N} + \frac{{|{\cal C}_{N-2}^{i_{N-2}}|}}{|{\cal C}_{N-1}^{i_{N-1}}|}\nonumber\\&(1+q_N)\biggl(q_{N-1} \tau_1 \cdots \tau_{N-1}+ \frac{{|{\cal C}_{N-3}^{i_{N-3}}|}}{|{\cal C}_{N-2}^{i_{N-2}}|}\biggl(\cdots\biggr) \biggr)\biggr)\Biggr\} \leq 0,
	\end{align}
	then the last term ${\cal J}\sum_{i}^{} \sum_{t_N = 0}^{\tau_{N}-1}\cdots \sum_{t_3=0}^{\tau_3-1}\sum_{t_2=0}^{\tau_2-1}\sum_{t_1=0}^{\tau_1-1}\\\mathbb{E}\left\{\|\nabla F(\mathbf{w}_{t_1,\ldots,t_{N}}^{i,t})\|^2\right\}$ is non-positive and can be removed from the right side of~\eqref{minus_bound}, and subsequently
	\begin{align}
		&\mathbb{E}\left\{F({\mathbf{w}}_{t+1})-F({\mathbf{w}}_{t})\right\} \leq -\frac{\mu}{2}\tau_N\cdots\tau_1 \mathbb{E}\left\{\|\nabla F( {\mathbf{w}}_{t})\|^2\right\}+\nonumber\\&\frac{L^2\mu^3}{2}\sigma^2 \tau_N \cdots \tau_2 \frac{\tau_1(\tau_1-1)}{2}+\frac{L^2\mu^3}{2N_\text{tot}} (1+q_1)\tau_N \cdots \tau_3\nonumber\\& \frac{\tau_2(\tau_2-1)}{2} \tau_1^2 \sigma^2  C_1+\frac{L^2\mu^3}{2N_\text{tot}}(1+q_2)(1+q_1)\tau_N \cdots \tau_4 \frac{\tau_3(\tau_3-1)}{2}\nonumber\\
		& \tau_2^2 \tau_1^2 \sigma^2C_2+\cdots+\frac{L^2\mu^3}{2N_\text{tot}}(1+q_{N-1})\cdots(1+q_1)\frac{\tau_N(\tau_N-1)}{2}\nonumber\\
		&\tau_{N-1}^2 \cdots \tau_1^2 \sigma^2C_{N-1}+\frac{L}{2}\mu^2(1+q_{N})\cdots (1+q_1)\tau_{N}\tau_{N-1}\cdots\nonumber\\
		&\tau_1\sigma^2  \frac{1}{N_\text{tot}}.
	\end{align}
	Then, by applying a telescoping sum over the global rounds $t \in \left\{0,\ldots,T-1\right\}$, we obtain
	\begin{align}
		\label{finnal_sttep}
		&\mathbb{E}\left\{F({\mathbf{w}}_{T})-F({\mathbf{w}}_{0})\right\} \leq -\frac{\mu}{2}\tau_N\cdots\tau_1 \sum_{t=0}^{T-1}\mathbb{E}\left\{\|\nabla F( {\mathbf{w}}_{t})\|^2\right\}\nonumber\\&+T\Biggl[\frac{L^2\mu^3}{2}\sigma^2 \tau_N \cdots \tau_2 \frac{\tau_1(\tau_1-1)}{2}+\frac{L^2\mu^3}{2N_\text{tot}} (1+q_1)\tau_N \cdots \tau_3\nonumber\\& \frac{\tau_2(\tau_2-1)}{2} \tau_1^2 \sigma^2  C_1+\frac{L^2\mu^3}{2N_\text{tot}}(1+q_2)(1+q_1)\tau_N \cdots \tau_4 \frac{\tau_3(\tau_3-1)}{2}\nonumber\\& \tau_2^2 \tau_1^2 \sigma^2C_2+\cdots+\frac{L^2\mu^3}{2N_\text{tot}}(1+q_{N-1})\cdots(1+q_1)\frac{\tau_N(\tau_N-1)}{2}\nonumber\\&\tau_{N-1}^2 \cdots \tau_1^2 \sigma^2C_{N-1}+\frac{L}{2}\mu^2(1+q_{N})\cdots\nonumber\\& (1+q_1)\tau_{N}\tau_{N-1}\cdots\tau_1\sigma^2  \frac{1}{N_\text{tot}}\Biggr],
	\end{align}
	and using the fact $\mathbb{E}\left\{F(\mathbf{w}_{T})\right\}\geq F^{*}$, we reach the conclusion of the proof.

\end{document}